\documentclass{article}

\usepackage{arxiv}

\usepackage[utf8]{inputenc} 
\usepackage[T1]{fontenc}    
\usepackage{hyperref}       
\usepackage{url}            
\usepackage{booktabs}       
\usepackage{amsfonts}       
\usepackage{nicefrac}       
\usepackage{microtype}      
\usepackage{graphicx}
\usepackage{doi}
\usepackage{natbib}

\usepackage{multirow}
\usepackage{amssymb}
\usepackage{amsmath}
\usepackage{tabularx}
\usepackage[ruled]{algorithm}
\usepackage[noend]{algpseudocode}
\usepackage{subcaption}
\usepackage{arydshln}
\usepackage[table]{xcolor}

\hypersetup{
	colorlinks=true,
	citecolor=blue,
	linkcolor=blue,
	urlcolor=blue
}

\title{BenDFM: A taxonomy and synthetic CAD dataset for manufacturability assessment in sheet metal bending}

\date{}

\newif\ifuniqueAffiliation
\uniqueAffiliationtrue

\ifuniqueAffiliation
\author{ \href{https://orcid.org/0009-0006-0881-5644}{\includegraphics[scale=0.06]{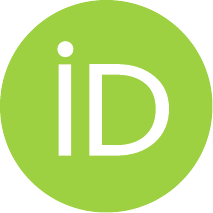}\hspace{1mm}Matteo Ballegeer}\thanks{Corresponding author} \\
	Ghent University\\
	CVAMO Core Lab\\
	Research Group Data Analytics \\
	\texttt{matteo.ballegeer@ugent.be} \\
	\And
	\href{https://orcid.org/0000-0001-9901-8507}{\includegraphics[scale=0.06]{orcid.pdf}\hspace{1mm}Dries F. Benoit} \\
	Ghent University\\
	CVAMO Core Lab\\
	Research Group Data Analytics \\
	\texttt{dries.benoit@ugent.be} \\
}
\else
\usepackage{authblk}

\newbox{\orcid}\sbox{\orcid}{\includegraphics[scale=0.06]{orcid.pdf}}
\author[1,2]{%
	\href{https://orcid.org/0009-0006-0881-5644}{\usebox{\orcid}\hspace{1mm}Matteo Ballegeer\thanks{\texttt{matteo.ballegeer@ugent.be}}}%
}
\author[1,2]{%
	\href{https://orcid.org/0000-0001-9901-8507}{\usebox{\orcid}\hspace{1mm}Dries F. Benoit\thanks{\texttt{dries.benoit@ugent.be}}}%
}
\affil[1]{Ghent University, Data Analytics Research Group, Faculty of Economics and Business Administration, Tweekerkenstraat 2, Belgium}
\affil[2]{FlandersMake@UGent---Corelab CVAMO, Tweekerkenstraat 2, Belgium}
\fi

\renewcommand{\headeright}{}
\renewcommand{\undertitle}{}
\renewcommand{\shorttitle}{}

\hypersetup{
pdftitle={BenDFM: A taxonomy and synthetic CAD dataset for manufacturability assessment in sheet metal bending},
pdfsubject={cs.LG},
pdfauthor={Matteo Ballegeer, Dries F. Benoit},
pdfkeywords={Design for manufacturing, Manufacturability taxonomy, Sheet metal bending, Synthetic CAD dataset, Geometric deep learning},
}

\begin{document}
\maketitle
\begin{abstract}
Predicting the manufacturability of CAD designs early, in terms of both feasibility and required effort, is a key goal of Design for Manufacturing (DFM).
Despite advances in deep learning for CAD and its widespread use in manufacturing process selection, learning-based approaches for predicting manufacturability within a specific process remain limited.
Two key challenges limit progress: inconsistency across prior work in how manufacturability is defined and consequently in the associated learning targets, and a scarcity of suitable datasets.
Existing labels vary significantly: they may reflect intrinsic design constraints or depend on specific manufacturing capabilities (such as available tools), and they range from discrete feasibility checks to continuous complexity measures.
Furthermore, industrial datasets typically contain only manufacturable parts, offering little signal for infeasible cases, while existing synthetic datasets focus on simple geometries and subtractive processes.
To address these gaps, we propose a taxonomy of manufacturability metrics along the axes of configuration dependence and measurement type, allowing clearer scoping of generalizability and learning objectives.
Next, we introduce BenDFM, the first synthetic dataset for manufacturability assessment in sheet metal bending.
BenDFM contains 20,000 parts, both manufacturable and unmanufacturable, generated with process-aware bending simulations, providing both folded and unfolded geometries and multiple manufacturability labels across the taxonomy, enabling systematic study of previously unexplored learning-based DFM challenges.
We benchmark two state-of-the-art 3D learning architectures on BenDFM, showing that graph-based representations that capture relationships between part surfaces achieve better accuracy, and that predicting metrics that depend on specific manufacturing setups remains more challenging.
Together, the taxonomy and BenDFM dataset provide a significant step toward deep-learning-assisted DFM in industrial design workflows.
\end{abstract}


\section{Introduction}\label{intro}
Design for Manufacturing (DFM) aims to identify production constraints early in the design phase to ensure manufacturability, reduce complexity and cost, and preserve functional intent~\citep{anderson2020design}. 
Early-stage validation has been shown to reduce prototyping cycles, lower production expenses, and accelerate time-to-market~\citep{sharp2018survey, wang2022machine}.
When manufacturability is not considered early, time and resources are often wasted, leading to compromised designs or, in extreme cases, the need to substantially revisit or restart the design process~\citep{patterson2021generation}.

Effective DFM involves two complementary stages~\citep{dewhurst1988early}. 
The first stage is selecting an appropriate manufacturing process.
The second is verifying manufacturability within the constraints of that chosen process.
We refer to these stages as process selection and intra-process manufacturability assessment (IPMA), respectively.
Process selection differentiates manufacturing processes using higher-level design attributes, while IPMA addresses subtler design differences within a single process that critically affect production feasibility and effort.

Both stages require detailed, process-specific knowledge that often lies beyond the expertise of designers.
In practice, decision-making is often divided across domains, with designers focusing on functional requirements and production engineers possessing deeper understanding of manufacturing constraints~\citep{patterson2021generation}.
This separation makes effective communication essential, yet coordination is frequently inadequate in real-world workflows~\citep{pullan2010application} and manufacturing knowledge is poorly disseminated and reused across organizations~\citep{favi2021cad}.
The absence of systematic methods for capturing and structuring DFM knowledge further leads to incomplete guidelines and reliance on individual experience~\citep{swift2013manufacturing}.
Interviews with design engineers in sheet metal bending have highlighted these challenges, revealing difficulties estimating design feasibility due to a lack of expertise, and that in practice, support tools are usually only available to the manufacturing department and not to the design engineer \citep{doellken2020challenges}.
Integrating DFM guidelines into CAD systems mitigates these challenges by enabling automated manufacturability analysis and guiding designers toward manufacturable and optimized designs from the earliest stages~\citep{ferrer2010methodology}.

Historically, automated DFM systems have relied on explicit knowledge encoding.
Traditional rule-based systems encode expert knowledge as explicit constraints~\citep{naranje2011knowledge, gupta2013classification}, but are rigid, scale poorly to novel geometries, and incur high maintenance overhead~\citep{favi2021cad}.
Moreover, design engineers who are expected to manually set the parameters for such rules do so by reviewing previous designs and internal design guidelines, and are often not aware of which parameters critically impact manufacturability~\citep{doellken2020challenges}.
Recent advances in deep learning have enabled data-driven alternatives that learn implicit manufacturing knowledge directly from data, reducing manual effort and improving adaptability~\citep{wang2022machine}.
However, the effectiveness of these methods depends on access to large, high-quality datasets~\citep{wang2018deep}.
Unfortunately, data scarcity remains a major barrier in manufacturing, particularly for CAD applications~\citep{yoo2021integrating} where intellectual property concerns over proprietary manufacturing knowledge further restrict availability~\citep{stjepandic2015intellectual}.
Consequently, research roadmaps have identified synthetic data generation as a key enabler for learning-based manufacturing systems~\citep{ordek2024machine, arinez2020artificial}.

Despite these obstacles, significant progress has been made in deep learning for manufacturing process selection, enabled by both industrial~\citep{yan2023automated, li2025classification} and synthetic datasets~\citep{zhao2024learning, hussong2025selection, zheng2025sfrgnn}.
In contrast, learning-based methods for IPMA remain largely unaddressed.
A fundamental limitation of industrial datasets in this context is their survivorship bias: they predominantly contain manufacturable, optimized designs, while unmanufacturable or suboptimal iterations are rarely archived~\citep{girsule2020data}.
This absence prevents deep learning models from learning the geometric features that cause production failures or increase manufacturing complexity, creating a critical blind spot for IPMA.

To address this imbalance, synthetic data generation offers a promising path forward, allowing controlled generation of both manufacturable and unmanufacturable designs. 
Yet, existing IPMA-focused synthetic datasets have significant limitations. 
They primarily feature simple, feature-augmented cubes tailored to subtractive manufacturing processes~\citep{ghadai2018learning, peddireddy2021identifying}, which have shown poor generalization to real-world CAD models~\citep{zheng2025sfrgnn, zhang2024brepmfr}.
Moreover, the dominance of subtractive processes represents a significant gap: while recent efforts have begun to address IPMA for more complex formative processes such as deep drawing~\citep{lehrer2025uscm}, no comparable datasets exist for sheet metal bending. 
Synthetically modeling sheet metal bending presents unique challenges: sequential operations must be captured accurately, and complex part-tooling interactions must be represented faithfully.

Beyond these data limitations, a more fundamental issue hampers progress: the inconsistent definition of ``manufacturability'' itself across the literature. 
Existing work varies along two critical dimensions, which must be clearly separated.
First, labels differ in how their validity depends on specific manufacturing capabilities. 
Some, such as manufacturable drilling hole depth-to-diameter ratios~\citep{ghadai2018learning}, are restricted by the available toolset, while others, like internal voids in milling~\citep{peddireddy2021identifying}, reflect geometric impossibilities and are universally valid within the process.
Second, the literature often conflates two distinct measures: \textit{feasibility} (whether a part can be produced) and \textit{complexity} (the effort, time, or cost of manufacturing). 
This lack of semantic clarity complicates the comparison of studies, hinders assessment of how well models and learned knowledge transfer across different manufacturing setups or companies, and ultimately impedes progress toward robust, deployable DFM systems.

This paper addresses these gaps with three main contributions.
First, we propose a taxonomy for manufacturability labels, structured along two dimensions: configuration dependence versus independence, and feasibility versus complexity.
This clarifies what manufacturability means, what scope each metric covers, and how well models trained on one type of label may transfer to new contexts or manufacturing setups.
Second, we introduce BenDFM, a synthetic dataset of 20,000 sheet metal parts generated through parametric modeling of sequential bending operations. 
The dataset captures complex part-tooling interactions and provides both folded and unfolded geometries, along with comprehensive manufacturability labels spanning all four taxonomy quadrants.
These labels range from tooling collisions to continuous metrics related to handling effort.
Third, we benchmark two state-of-the-art 3D deep learning models on BenDFM, showing that topology-aware representations are essential for accurate predictions, and that configuration-dependent metrics remain challenging to learn.
Together, this study advances learning-based manufacturability assessment by clarifying the concept through a comprehensive taxonomy, operationalizing it with a novel synthetic dataset for sheet metal bending, and validating both through benchmarking.

The remainder of this paper is organized as follows: Section~\ref{sec:related_work} reviews relevant literature; Section~\ref{sec:taxonomy} introduces our manufacturability taxonomy; Section~\ref{sec:data_generator} describes the BenDFM dataset and its generation process; Section~\ref{sec:benchmark} presents benchmarking experiments and results; Section~\ref{sec:discussion} discusses implications, limitations, and future directions; and Section~\ref{sec:conclusion} concludes the study.

\section{Related work}\label{sec:related_work}

This section reviews the evolution of automated DFM systems, highlighting the shift from rule-based to data-driven approaches.
We highlight the uneven progress between process selection and IPMA, underscoring the unique data-related challenges that have limited advances in the latter.
Finally, we show how manufacturability labels in current IPMA studies differ widely in definition, generalizability, and utility for deep learning, contributing to an ambiguous research landscape.

\subsection{Manufacturing process selection}

Manufacturing process selection is the task of determining the most appropriate manufacturing process for a given design~\citep{swift2013manufacturing}. 
Traditionally, this has relied on expert judgment or rule-based systems that required extensive user input~\citep{hoefer2018automated}. 
The advent of data-driven approaches has enabled automated analysis of 3D CAD models through (deep) learning. 
By learning from geometric patterns, models can determine the most suitable manufacturing process for a given design.
These methods operate at two distinct levels: (1) whole-part classification, where an entire design is assigned to a single process, and (2) segmentation, where regions of a part are mapped to different processes. 

A challenge in using deep learning in this setting is that native CAD boundary representations (B-reps) are not directly compatible with standard neural network architectures. 
Consequently, the input representation must be carefully chosen, as this decision drives the architecture and performance of downstream models. 
Early studies bypassed direct 3D analysis, relying instead on simplified descriptors or 2D projections. 
For example,~\cite{hamouche2018classification} used curvature descriptors with 2D Convolutional Neural Networks (CNNs) to classify parts into six forming processes, while others extracted part-level geometric features for decision-tree-based classification~\citep{zhao2020automated} or used heat kernel signatures with 2D CNNs~\citep{wang2023manufacturing}.

Recent advances have enabled direct learning on 3D representations, a shift accelerated in subtractive manufacturing by a close link to the more established field of \textit{manufacturing feature recognition}.
Since features like holes and pockets often map directly to operations like drilling or milling, methods and datasets from feature recognition were readily adapted for process selection.
The conceptual similarity explains the prevalence of synthetic datasets built on a common methodology: parametrically extruding subtractive features to simple base solids (typically cubes), as seen in prominent feature recognition datasets like FeatureNet~\citep{featurenet}, MFCAD~\citep{mfcad}, and MFCAD++~\citep{colligan2022hierarchical}.

This move to 3D spurred the adoption of various representations from the broader geometric deep learning community. 
Voxel-based 3D CNNs were applied for both process classification~\citep{zhao2024learning} and feature recognition~\citep{ning2023part}.
Point cloud methods, operating on points sampled from the model surface, have been used for feature segmentation~\citep{zhang2022machining, zhang2024point} and process selection across multiple domains~\citep{wang2023manufacturing2, liu2025manufacturing}.
More recently, Graph Neural Networks (GNNs) operating on the B-rep's Attributed Adjacency Graph (AAG) have become prominent for manufacturing applications. 
In this representation, faces become graph nodes, their adjacencies become edges, and geometric properties become attributes. 
This structure preserves both the topology and geometry of the original B-rep without the information loss associated with other discretization methods.

Beyond supervised methods, research has explored unsupervised learning methods, such as using autoencoders to learn latent process capabilities for milling and turning~\citep{yan2023automated, yan2024deep}. 
Others have incorporated a temporal dimension by predicting sequences of machining operations~\citep{zhao2022data, zhao2024deep, xiao2025novel}. 
Overall, the process selection field has seen rapid progress, benefiting from advances in deep learning and the availability of large-scale synthetic datasets in the related field of feature recognition.

\subsection{Intra-process manufacturability assessment}

Intra-process manufacturability assessment (IPMA) aims to evaluate whether a design can actually be produced within the constraints of a selected process~\citep{dewhurst1988early}. 
Rather than discriminating between processes based on key design characteristics, IPMA must capture the often subtle and complex within-process differences among geometrically similar parts that determine manufacturability. 
This, in turn, requires precise modeling of process capability boundaries such as tool accessibility, collision avoidance, and other geometric constraints.

Traditionally, IPMA systems have also relied on rule-based approaches.
These include knowledge-based tools for deep drawing~\citep{naranje2011knowledge}, extensive rule sets for sheet metal bending~\citep{kumar2016feature}, and computational metrics for accessibility, tool requirements, and machine axis count in milling~\citep{hoefer2018automated, yeo2021manufacturability, chen2021design}. 
While the process selection literature has rapidly embraced deep learning, its adoption in IPMA has been limited.
A major barrier is the scarcity of documented failure cases in industrial datasets~\citep{girsule2020data}, which are essential for learning the geometric factors underlying unmanufacturability.
A notable exception is the study by~\cite{seibold2022process}, who used 2D projections of 3D sheet-metal parts and CNNs to classify coatability, leveraging a dataset collected over time that explicitly preserved uncoatable cases.
While industrial datasets have further proven useful for cost estimation of successful designs~\citep{ning2020manufacturing, ning2020manufacturing2, zhang2022novel}, their lack of failure cases has compelled IPMA researchers to turn to synthetic data generation.

This has produced several noteworthy IPMA datasets, yet these efforts reveal a set of shared limitations.
The earliest examples focused on subtractive processes using simple, cube-based geometries.
For instance,~\cite{ghadai2018learning} used 3D CNNs to assess the manufacturability of drilled holes based on depth-to-diameter ratios, while~\cite{peddireddy2021identifying} identified non-manufacturable features like voids and undercuts in milled and turned parts.
While foundational, the reliance on primitive geometries has been shown to limit model generalization to real-world CAD parts~\citep{zheng2025sfrgnn,zhang2024brepmfr}.
More recent work has applied simulation-based labeling to existing (non-DFM) datasets.
For example,~\cite{zhong2025deepmill} used a collision detection algorithm to generate tool accessibility labels for milling, and ~\cite{chen2025virl} applied CAM simulations to extract complexity metrics like machining time and blade interference hazards.
These studies mark a step toward greater realism, but as summarized in Table~\ref{tab:synthetic_datasets}, the landscape of synthetic data for IPMA remains heavily skewed toward subtractive processes, with a clear lack of datasets designed from the ground up for complex formative manufacturing domains.

\begin{table}[ht]
	\centering
	\small
	\setlength{\tabcolsep}{3pt}
	\renewcommand{\arraystretch}{0.65}
	\caption{Overview of introduced synthetic DFM datasets.}
	\begin{tabularx}{\linewidth}{p{3.8cm} >{\centering\arraybackslash}p{2cm} >{\centering\arraybackslash}p{2cm} >{\centering\arraybackslash}p{1.6cm} >{\centering\arraybackslash}p{1.6cm} >{\centering\arraybackslash}p{1.4cm} >{\centering\arraybackslash}p{1.4cm}}
		\toprule
		\multirow{3}{*}{\textbf{Study}} & \multicolumn{2}{c}{\textbf{Process Type}} & \multicolumn{2}{c}{\textbf{Base Design}} & \multicolumn{2}{c}{\textbf{DFM Stage}} \\
		\cmidrule(lr){2-3} \cmidrule(lr){4-5} \cmidrule(l){6-7}
		& \textbf{Subtractive} & \textbf{Formative} & \textbf{Primitive} & \textbf{Complex} & \textbf{Selection} & \textbf{IPMA} \\
		\midrule
		\cite{zhao2024learning} & \checkmark &  & \checkmark &  & \checkmark &   \\
		\cite{hussong2025selection} & \checkmark &  & \checkmark &  & \checkmark &   \\
		\cite{zheng2025sfrgnn} & \checkmark &  & \checkmark &  & \checkmark &   \\
		\cite{featurenet} & \checkmark &  & \checkmark &  & \checkmark &   \\
		\cite{mfcad} & \checkmark &  & \checkmark &  & \checkmark &   \\
		\cite{colligan2022hierarchical} & \checkmark &  & \checkmark &  & \checkmark & \\
		\cite{ning2023part} & \checkmark &  & \checkmark &  & \checkmark &   \\
		\cite{zhang2022machining} & \checkmark &  & \checkmark &  & \checkmark &   \\
		\cite{ghadai2018learning} & \checkmark &  & \checkmark &  &  & \checkmark \\
		\cite{peddireddy2021identifying} & \checkmark &  & \checkmark &  &  & \checkmark \\
		\cite{zhong2025deepmill} & \checkmark &  &  & \checkmark* &  & \checkmark \\
		\cite{chen2025virl} & \checkmark & &  & \checkmark* &  & \checkmark \\
		\cite{wang2023manufacturing} & \checkmark & molding &  & \checkmark & \checkmark &  \\
		\cite{lehrer2025uscm} &  & deep drawing &  & \checkmark & & \checkmark  \\
		\midrule
		\textbf{BenDFM} &  & bending &  & \checkmark &  & \checkmark \\
		\bottomrule
	\end{tabularx}
	\caption*{\raggedright * Label existing non-DFM datasets via simulation software instead of synthesizing new designs.}
	\label{tab:synthetic_datasets}
\end{table}

Only a handful of recent studies address processes and geometries of greater complexity.
Concurrent with this work,~\cite{lehrer2025uscm} introduced a synthetic dataset for deep-drawn U-channel components that includes drawability labels. 
Other studies have examined sheet metal bending, but none have yet produced datasets suitable for IPMA.
The SMCAD dataset~\citep{smcad} advances feature recognition by incorporating complex features (e.g., hinges, stiffeners, ear plates) into a limited set of predefined synthetic base geometries. 
While this approach moves beyond primitive shapes, its small, fixed set of base geometries means the variation in designs comes from added features (e.g., holes, hinges) rather than from diverse bending flanges, limiting its applicability to IPMA, where manufacturability depends heavily on geometric interactions between bends.
\cite{barda2023generative} proposed a generative framework that connects a load to floor contact regions with a sheet metal structure while minimizing material use and satisfying constraints such as avoiding collisions with the environment and flat-pattern overlaps. 
Although this represents a promising approach to constrained shape generation, the work does not provide a dataset or DFM labels, limiting its relevance for deep learning-based IPMA.

The need for a large-scale, publicly available dataset specifically designed for sheet metal bending is evident.
Moreover, its importance extends beyond the IPMA context, as prior research in related areas, such as bending feature recognition, has relied on a highly constrained set of benchmark parts (typically 3-40 designs) reused across multiple studies~\citep{gupta2013classification, kumar2016feature, salem2017towards}.
This lack of diverse, representative data highlights a critical barrier to robust validation and methodological advancement.
We address this gap with BenDFM, a large-scale dataset of geometrically diverse synthetic bending parts, generated through an advanced, process-aware pipeline that provides rich manufacturability labels and metadata.

\subsection{Ambiguity in defining manufacturability}

The limited body of research on data-driven IPMA exposes a more fundamental challenge: the term ``manufacturability'' is inconsistently defined throughout the DFM literature. 
A central source of this inconsistency is the varying degree to which manufacturability labels depend on specific tooling, machine capabilities, and operational parameters. 
This lack of semantic clarity makes it difficult to assess the generalizability of deep learning solutions for DFM, and to compare studies that use different definitions of manufacturability.

For example, consider IPMA studies treating manufacturability as binary feasibility.
\cite{ghadai2018learning} label drilled holes as unmanufacturable based on depth-to-diameter ratio thresholds.
\cite{peddireddy2021identifying} flag undercuts as unmanufacturable under 3-axis milling assumptions, and \cite{zhong2025deepmill} assign binary collision labels based on accessibility with a specific ball-end cutter.
In each case, manufacturability is highly configuration-dependent, as these binary labels could change with a different drilling setup, a 5-axis milling machine, or an alternative cutter geometry, respectively.
On the other hand,~\cite{peddireddy2021identifying} also flag internal voids as unmanufacturable, which can be regarded as a geometric impossibility in subtractive manufacturing that is configuration-independent.

Similarly, for continuous labels, the extent of configuration dependency varies. 
\cite{hoefer2018automated} and~\cite{yeo2021manufacturability} propose part-level complexity metrics (e.g., required rotations, deviation angles) that depend solely on geometry and are expected to generalize across environments within a given process. 
In contrast,~\cite{chen2025virl} and~\cite{lehrer2025uscm} use CAM simulation software to provide manufacturability labels, coupling their validity to specific simulation parameters used when creating the labels. 
Likewise, the predictions of manufacturing cost estimation models~\citep{ning2020manufacturing, ning2020manufacturing2, zhang2022novel} are intrinsically linked to the specific production environment from which the training data was sourced.

Although several taxonomies of manufacturability have been proposed, they do not systematically address configuration dependency, which we argue is central to data-driven DFM systems.
For example,~\cite{shankar1993generalized} distinguished between generalized and domain-specific manufacturability using compatibility, complexity, and coupling indices.
\cite{gupta1997automated} classified labels as binary, qualitative, abstract quantitative, or cost/time-based. 
\cite{hoefer2018automated} categorized metrics by geometric scope (aggregate, face-based, or slice-based), while \cite{yeo2021manufacturability} distinguished between tool-based criteria (e.g., depth) and shape-based criteria (e.g., symmetry).

By not explicitly considering configuration dependency, these taxonomies overlook a critical dimension that influences model generalizability and interpretability.
Consequently, researchers cannot reliably evaluate and compare different learning-based approaches, and practitioners cannot confidently deploy them across varying production setups. 
To address this fundamental gap, we propose a revised taxonomy that integrates a data-driven perspective on manufacturability and clearly distinguishes between different classes of manufacturability metrics.

\section{A taxonomy for manufacturability metrics}\label{sec:taxonomy}

\begin{figure}[htp]
	\centering
	\includegraphics[width=0.8\textwidth]{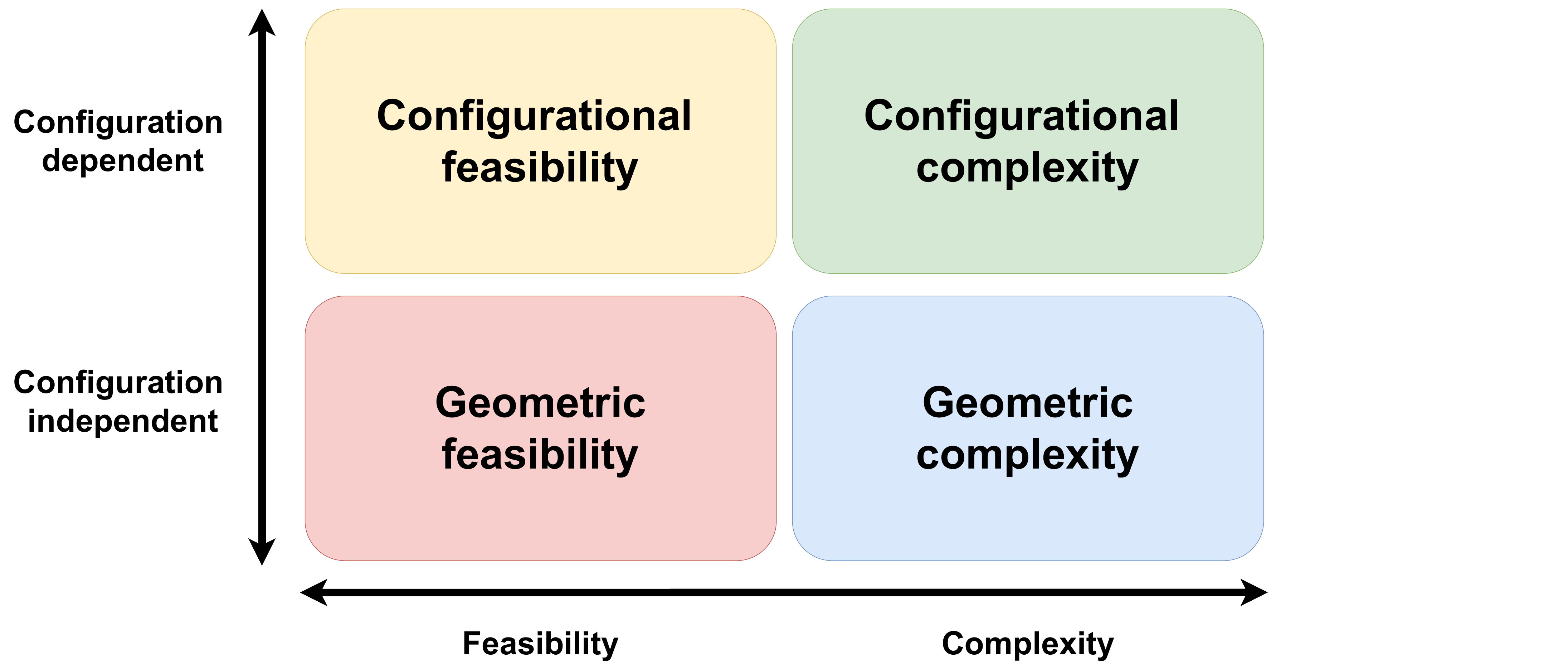}
	\caption{Proposed taxonomy for manufacturability metrics.
	Each quadrant represents a distinct class of manufacturability evaluation relevant for data-driven DFM.}\label{fig:taxonomy}
	\end{figure}

To bring clarity to the term ``manufacturability'' and its use in data-driven models, we propose a taxonomy that categorizes manufacturability metrics along two orthogonal axes.
The first axis defines the extent to which a metric is tied to a specific production environment. 
It ranges from \textit{configuration-independent} metrics (determined purely by part geometry, valid regardless of tooling or machine capabilities) to \textit{configuration-dependent} metrics (contingent on specific machine capabilities, tooling, and operational parameters that may vary across different manufacturing environments).
The second axis distinguishes between whether a metric evaluates a design's feasibility versus the effort required to produce it.
\textit{Feasibility} metrics answer the question ``Can this part be made?'' by serving as discrete checks on producibility, typically by identifying violations of hard constraints. 
These are often translated into binary metrics (e.g., manufacturable/unmanufacturable)~\citep{ghadai2018learning}, but could also be discrete counts of violations to reflect violation severity.
\textit{Complexity} metrics, in contrast, answer the question ``How difficult is it to make?'' by providing continuous measures of manufacturing effort, such as time, cost, or abstract indicators of difficulty.
As illustrated in Figure~\ref{fig:taxonomy}, the intersection of these two axes defines four distinct quadrants, each representing a unique category of manufacturability assessment with specific data requirements, learning objectives, and implications for the generalizability of predictive models.

\subsection{Geometric feasibility}

Geometric feasibility captures hard constraints rooted in part geometry and basic physical laws, independent of any specific manufacturing setup. 
These are universally prohibitive within the considered manufacturing process: no machine, tool, or different sequence of operations can overcome them.
Examples include internal voids that are unreachable in subtractive manufacturing~\citep{peddireddy2021identifying} or self-intersections in the unfolded state of a sheet metal part~\citep{barda2023generative}.
Because of this, models trained on such metrics are highly generalizable and particularly useful in early-stage design validation. 
However, their scope is narrow: they flag only severe design flaws and miss the nuanced, process-dependent realities that define manufacturability in industrial contexts. 
Their main value lies in universality and interpretability, not industrial specificity.

\subsection{Configurational feasibility}

Configurational feasibility assesses whether a design is manufacturable within a specific production environment given a defined set of machines, tools, operational parameters and sequence. 
A part's feasibility may vary between setups depending on machine axes, tool reach, or fixture constraints.
Examples from the literature include flagging drilling holes as unmanufacturable based on prespecified depth-to-diameter ratios~\citep{ghadai2018learning} and assigning binary collision labels based on accessibility with a specific cutter~\citep{zhong2025deepmill}.
Such metrics are often derived from simulations, CAM toolchains, or expert annotation, and can be highly relevant for deployment in matching environments. 
However, their dependency on specific configurations can limit model transferability across different production settings.
For systems operating solely on CAD input (without configuration knowledge), it is crucial to recognize the assumptions baked into such metrics. 
Including configuration descriptors as model inputs could improve generalizability, but such information may not be available during early design. 

\subsection{Geometric complexity}

Geometric complexity metrics describe continuous, geometry-intrinsic indicators of manufacturing difficulty.
These measures are defined purely by the CAD geometry, enabling broad applicability and comparative design optimization.
Examples from the literature include part-level metrics like sharpness (minimum internal corner radius), drilling hole depth-to-diameter ratios, and the number of angles required to reach every part of the design in milling~\citep{hoefer2018automated, yeo2021manufacturability}.
It is important to note that metrics like the drilling hole depth-to-diameter ratio depend exclusively on the part's geometry and are classified as geometric complexity metrics for this reason.
However, when a threshold is applied to define acceptable values~\citep{ghadai2018learning}, this introduces assumptions about specific tooling capabilities, converting the metric into a configurational feasibility criterion instead.
While generalizable, geometric complexity metrics do not account for configuration-specific factors (e.g., machine speed, available tool set, tool change time) that can dominate real-world costs and manufacturing effort. 
In the context of this taxonomy, geometric complexity is treated as a training label; nevertheless, such part-level metrics are also commonly used as input shape descriptors for downstream tasks such as process selection~\citep{hamouche2018classification, zhao2020automated}.

\subsection{Configurational complexity}

Configurational complexity quantifies setup-specific, continuous measures of manufacturing effort. 
These metrics can take abstract forms (e.g., number of tool changes), time-based values (e.g., cycle or setup time), or cost-related indicators, which often correlate closely. 
Typical data sources include CAM simulations and historical industrial records. 
Examples from literature are studies using manufacturing cost~\citep{ning2020manufacturing, ning2020manufacturing2, zhang2022novel}, simulation-derived metrics like accessibility with prespecified tool geometries~\citep{zhong2025deepmill} or machining time~\citep{chen2025virl}.
Because these labels directly reflect operational realities, they can provide highly actionable insights for optimization within particular manufacturing environments. 
However, this specificity limits their generalizability: models trained to predict configurational complexity might not transfer well to environments with vastly different capabilities.
 
\subsection{Implications, challenges and opportunities for DFM systems}

The proposed taxonomy reveals fundamental trade-offs in the design of data-driven DFM systems.
Geometric labels offer generalizability and are suited to early-stage design. 
However, they often lack the specificity needed for practical deployment in real manufacturing environments.
Configurational labels offer industrial relevance and precision, but require tightly scoped training and deployment contexts to avoid poor generalization.
These trade-offs are especially important in early-stage DFM tools, where production details may be unavailable. 
In such cases, geometric complexity offers a promising middle ground, delivering actionable feedback without the need for full context. 
Metrics such as the required machine axes, part symmetry, tolerance density, or PMI elements, potentially aggregated into a composite manufacturability score, can support both design evaluation and model interpretability. 
Alternatively, models trained on configuration-dependent inputs could output confidence scores based on configurational similarity between training and deployment contexts if this information is known.

Each quadrant also presents unique data challenges.
In industrial datasets, the most common manufacturability metrics are those for configurational complexity, particularly those relating to manufacturing time and cost~\citep{ning2020manufacturing, ning2020manufacturing2, zhang2022novel}. 
However, this data is subject to survivorship bias: because industrial designs are typically optimized for low manufacturing effort, models trained on them might make overly optimistic predictions for new, suboptimal designs. 
For feasibility metrics, this survivorship bias manifests as a complete absence of negative examples, since unmanufacturable designs are rarely retained in industrial databases~\citep{girsule2020data}.
This lack of failure data makes synthetic generation essential for training models to recognize configuration-dependent infeasibilities, such as tooling collisions.
Geometric infeasibilities are even less likely to appear in industrial datasets, yet their more universal nature is expected to make synthetic insights more generalizable.
While we define the generalizability of data-driven models by the characteristics of their training labels, this reasoning extends equally to unsupervised methods such as those by~\cite{yan2023automated,yan2024deep} that learn latent spaces directly from CAD geometries. 
The degree of configuration-specific information encoded in these latent spaces will ultimately determine the generalizability and applicability of the learned representations to new designs.

Ultimately, this taxonomy highlights the strengths and limitations of models trained on labels from each quadrant, and it underscores the potential need for hybrid systems that integrate geometric generalizability with configurational specificity.
By categorizing DFM learning targets from a deep learning perspective, it helps researchers and practitioners align their terminology, clarify objectives, and explicitly define deployment assumptions.
Such conceptual clarity is critical for developing DFM systems that are both robust and practically deployable.

\section{The BenDFM dataset}\label{sec:data_generator}

This section introduces the BenDFM dataset, detailing how designs are generated and how the dataset incorporates manufacturability labels across our proposed taxonomy.
Dataset creation is based on emulating sheet metal bending operations through parametric CAD modeling using PythonOCC~\citep{paviot2022pythonocc}.
The generation process is guided by two main objectives. 
The first is to produce a geometrically diverse set of designs that represent the practical design space in sheet metal bending. 
The second is to provide accurate manufacturability labels, including both feasible and infeasible designs, that span the four quadrants of our taxonomy. 
Geometric diversity is achieved by stochastic sampling of parameters such as bend angles, radii, and bend sequences.
Manufacturability labeling is accomplished by logging geometric metrics and simulating bend operations with parametrically defined tooling.
This enables detection of collisions and quantification of part reorientation requirements.
The resulting dataset comprises a wide variety of 3D bent parts, their corresponding flat patterns, complete bend sequences with associated parameters, and a rich set of manufacturability labels covering all four taxonomy quadrants.

\subsection{Parametric bend generation}\label{sec:bend_simulation}

\begin{figure}[htp]
	\centering

	\begin{subfigure}[b]{0.20\textwidth}
		\includegraphics[width=\textwidth]{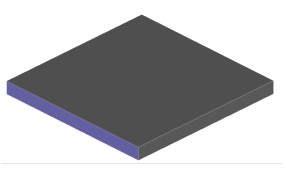}
		\caption{Select bend edge}\label{fig:bend_creation_a}
	\end{subfigure}
	\hfill
	\begin{subfigure}[b]{0.17\textwidth}
		\includegraphics[width=\textwidth]{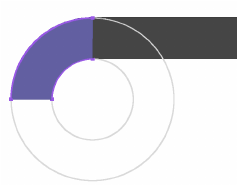}
		\caption{Build 2D face}\label{fig:bend_creation_b}
	\end{subfigure}
	\hfill
	\begin{subfigure}[b]{0.20\textwidth}
		\includegraphics[width=\textwidth]{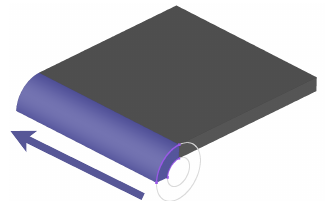}
		\caption{Extrude bend}\label{fig:bend_creation_c}
	\end{subfigure}
	\hfill
	\begin{subfigure}[b]{0.17\textwidth}
		\includegraphics[width=\textwidth]{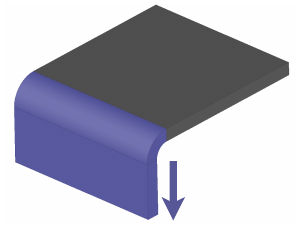}
		\caption{Extrude flange}\label{fig:bend_creation_d}
	\end{subfigure}
	\hfill
	\begin{subfigure}[b]{0.20\textwidth}
		\includegraphics[width=\textwidth]{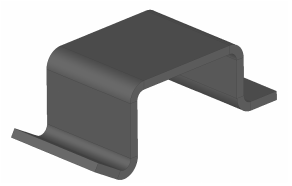}
		\caption{Repeat}\label{fig:bend_creation_e}
	\end{subfigure}
	\caption{Core steps of the bend creation process.}\label{fig:bend_creations}
\end{figure}

The core data generation process simulates sheet metal bending by incrementally adding flanges to an initial planar sheet. 
This sheet is defined by configurable parameters (length, width, and thickness) and can be oriented in any of the three principal planes (XY, YZ, or XZ).
The flange generation process consists of a structured sequence of operations designed to simulate realistic sheet metal bending while maintaining geometric diversity. 
The procedure maintains a mapping of eligible edges on each face, enabling tracking of bendable edges as the part evolves. 
Each bend is added through a five-step procedure, as illustrated in Figure~\ref{fig:bend_creations}.

First, an eligible edge is selected from the pool of candidates using a weighted sampling strategy. 
Edges closer to the base flange and those with greater length are prioritized (Figure~\ref{fig:bend_creation_a}). 
This encourages designs with denser bends near the base and discourages the formation of long snaking patterns with parallel bends extending far from the base. 
Favoring longer edges also mitigates the overuse of short flange sides, avoiding excessively narrow, often unrealistic flanges.

A 2D bend face is then constructed based on sampled bend parameters (Figure~\ref{fig:bend_creation_b}). 
These include bend angle, radius, orientation (up or down), and sheet thickness, drawn from discrete and continuous distributions reflecting manufacturing norms.
For instance, bend angles are selected from a set of common values, with $90^\circ$ bends favored.
Starting from the corner of the selected edge, the bend radius and orientation (up or down) are used to determine the bend center, which, together with the sheet thickness, defines the inner and outer circles that characterize the bend. 

The 3D bend shape is created by extruding the 2D bend face from the previous step along the direction of the bend edge (Figure~\ref{fig:bend_creation_c}), while the new bend flange is produced by extruding the long face of the newly created bend to the specified flange height (Figure~\ref{fig:bend_creation_d}).
Following each bend, the system updates all geometric metadata. 
The newly bent edge is removed from the candidate pool, the new flange's edges are added, and the operation is logged with its parameters. 
This process repeats until a target number of bends is reached (Figure~\ref{fig:bend_creation_e}).

\begin{figure}[htp]
	\centering
	\begin{subfigure}[b]{0.27\textwidth}
		\includegraphics[width=\textwidth]{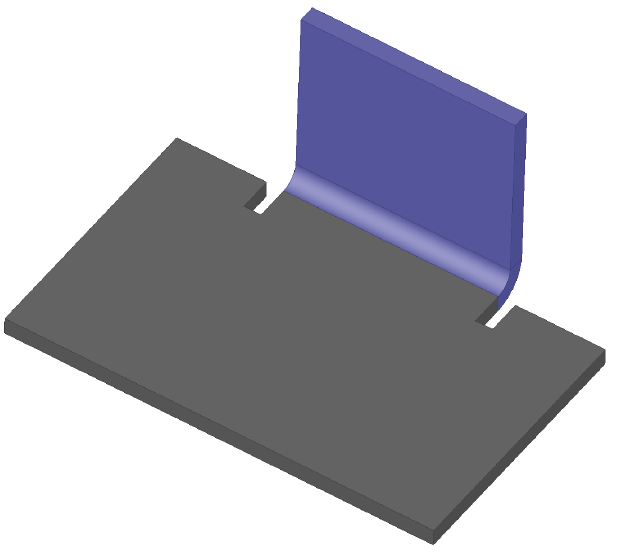}
		\caption{Bend reliefs}\label{fig:bend_reliefs}
	\end{subfigure}
	\hspace{0.05\textwidth}
	\begin{subfigure}[b]{0.27\textwidth}
		\includegraphics[width=\textwidth]{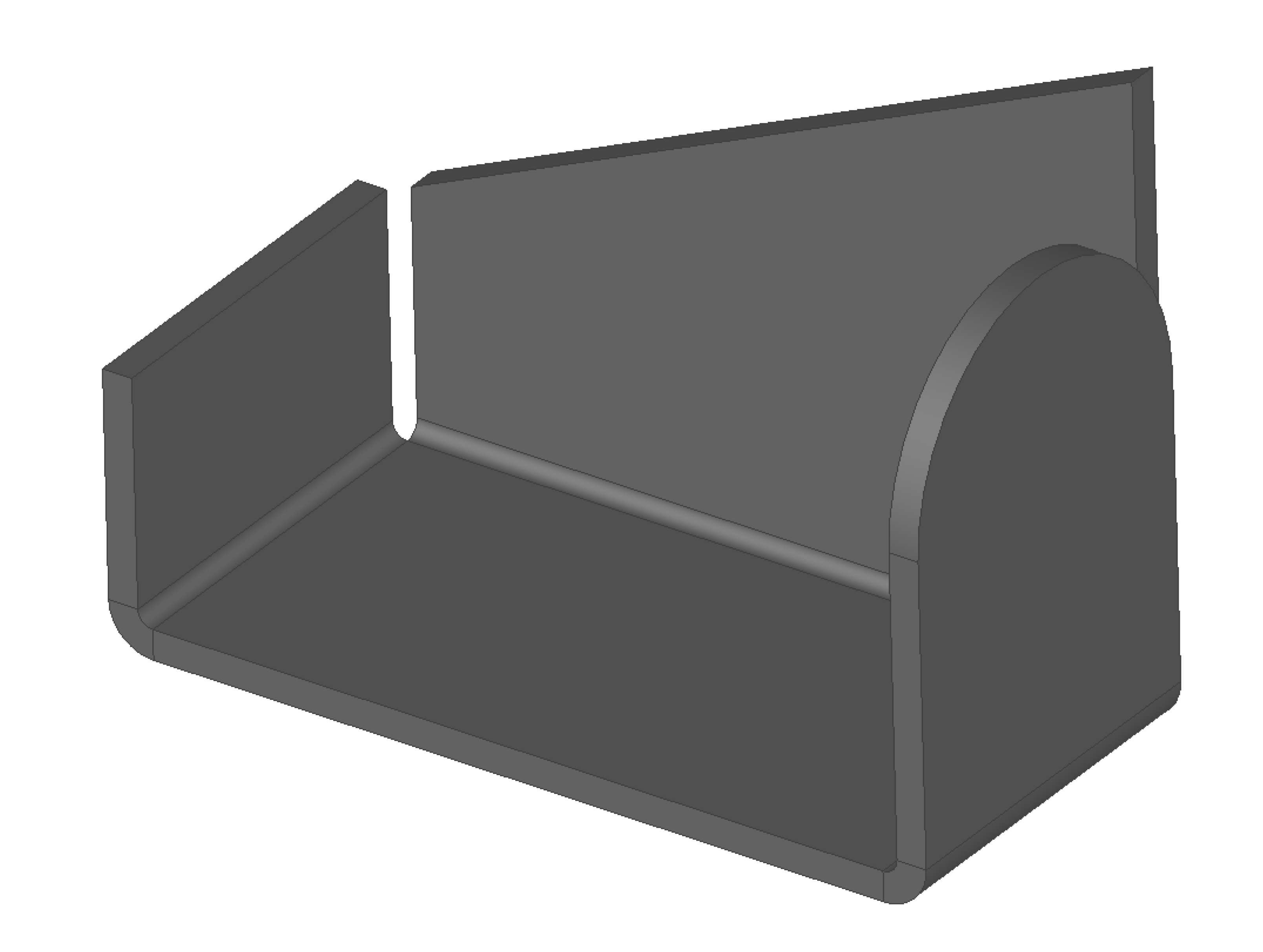}
		\caption{Flange geometry variants}\label{fig:flange_types}
	\end{subfigure}
	\hspace{0.05\textwidth}
	\begin{subfigure}[b]{0.24\textwidth}
		\includegraphics[width=\textwidth]{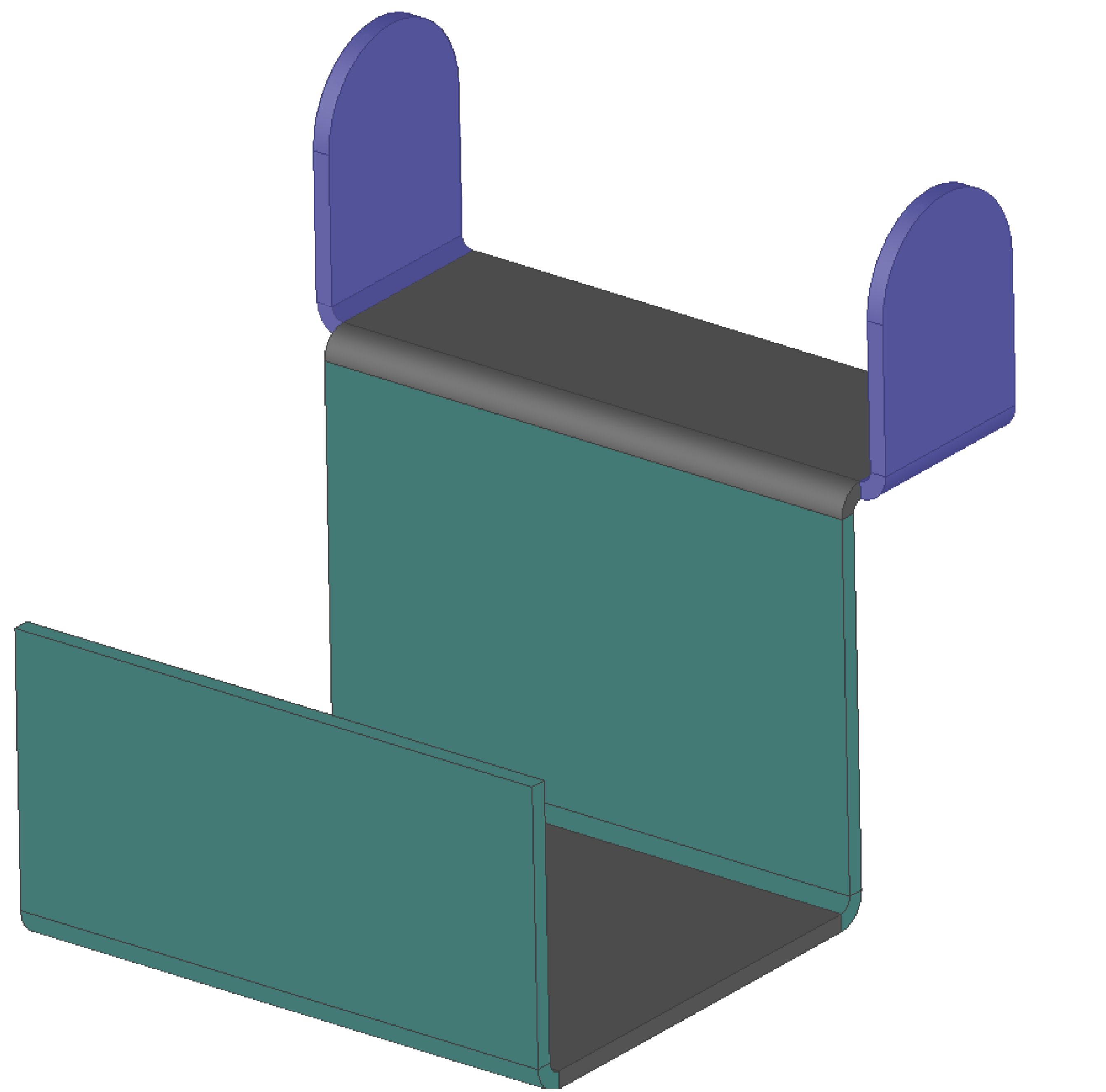}
		\caption{Symmetry bias}\label{fig:symmetry}
	\end{subfigure}
	\caption{Bend reliefs, flange types, and symmetry bias enhance the realism and variability of generated designs.}
\end{figure}

Three additional design considerations are incorporated to improve the realism of the generated parts.
\begin{itemize}
 	\item \textbf{Bend reliefs}.
	For a subset of bends, the bend width is sampled to be between 50--75\% of the edge length. 
	In such cases, bend reliefs are inserted to prevent tearing and distortion (Figure~\ref{fig:bend_reliefs}).
	\item \textbf{Flange geometry variants}.
	Flanges are generated in three forms: rectangular, slanted, and rounded (Figure~\ref{fig:flange_types}). 
	The flange shape is chosen during extrusion (Step 4), with rectangular and slanted flanges being more commonly generated and rounded flanges serving as terminal flanges (they do not produce new eligible edges for bends).
	\item \textbf{Symmetry bias}.
	To reflect common symmetrical properties of real-world bending parts, the generator is biased to create symmetric flange pairs. 
	After a successful bend addition, for a proportion of bends, the algorithm checks for a symmetric edge counterpart on the same face. 
	If found, the bend parameters from the preceding bend are reused on this edge while permitting variations in flange height, as illustrated in Figure~\ref{fig:symmetry}.
\end{itemize}

\subsection{Tooling geometry modeling}

\begin{figure}[htp]
	\centering
	\begin{subfigure}[b]{0.3\textwidth}
		\includegraphics[width=\textwidth]{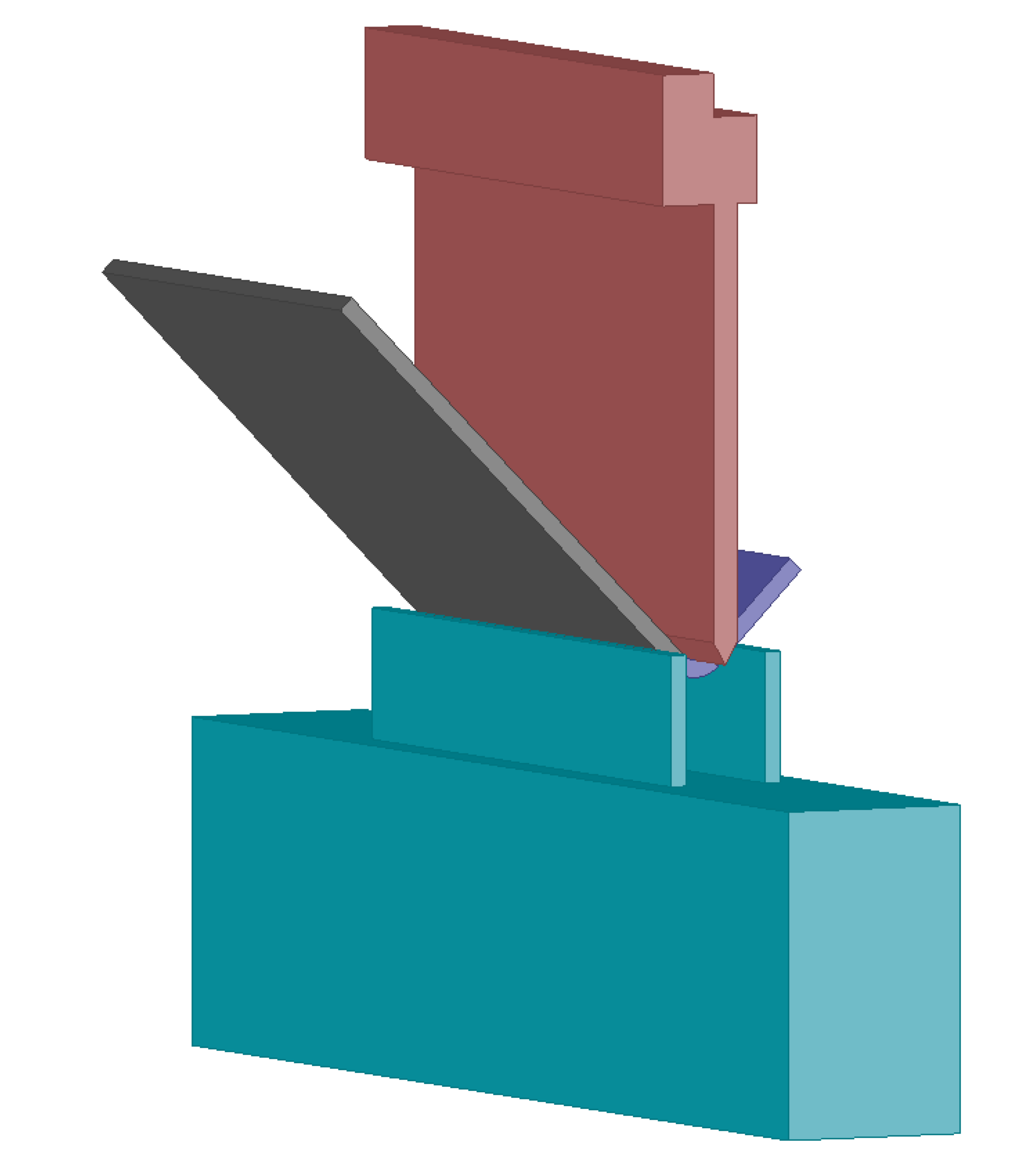}
		\caption{Punch and die geometry}\label{fig:punch_and_die}
	\end{subfigure}
	\hfill
	\centering
	\begin{subfigure}[b]{0.3\textwidth}
		\includegraphics[width=\textwidth]{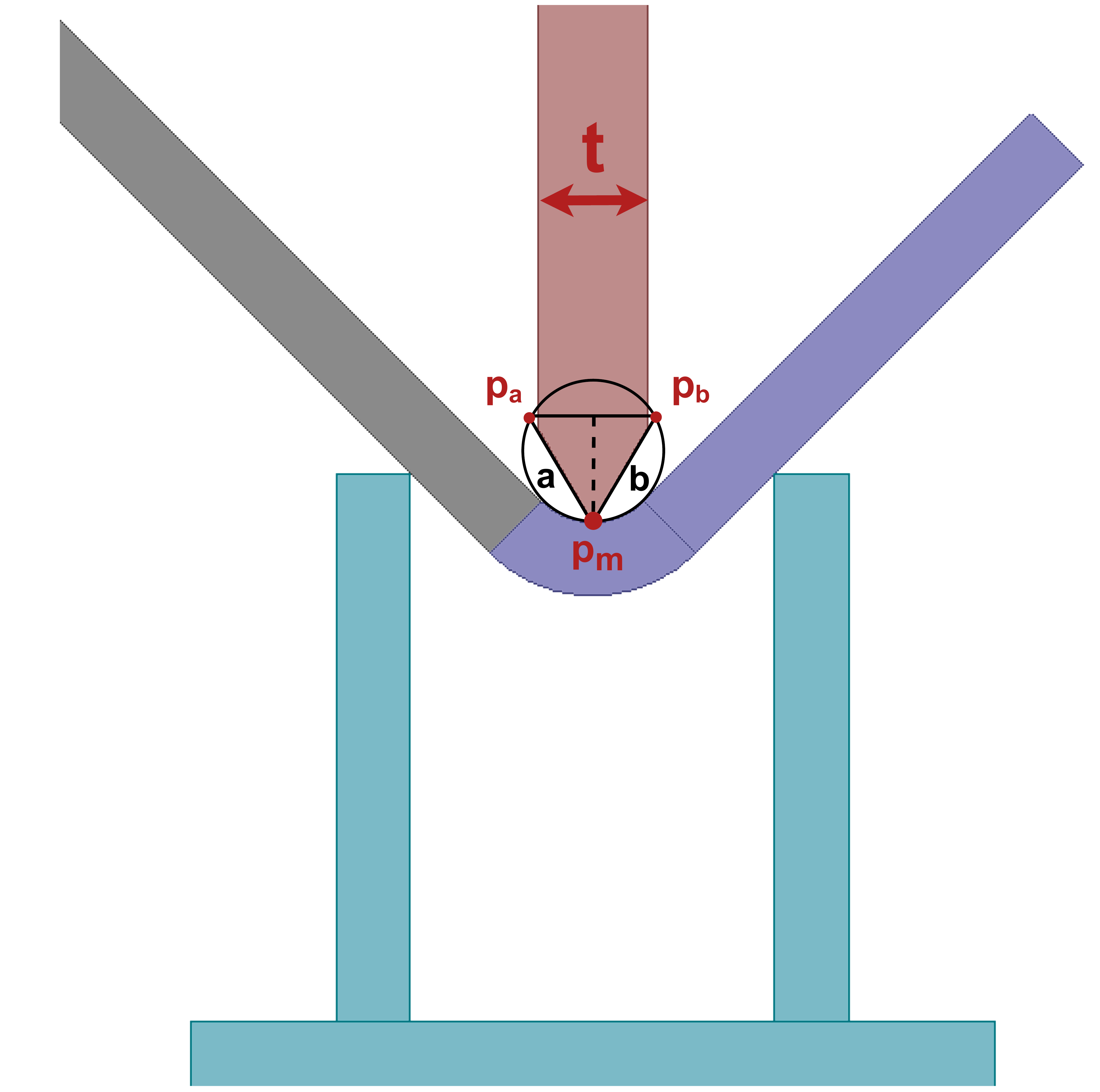}
		\caption{Punch construction}\label{fig:punch_creation}
	\end{subfigure}
	\hfill
	\begin{subfigure}[b]{0.3\textwidth}
		\includegraphics[width=\textwidth]{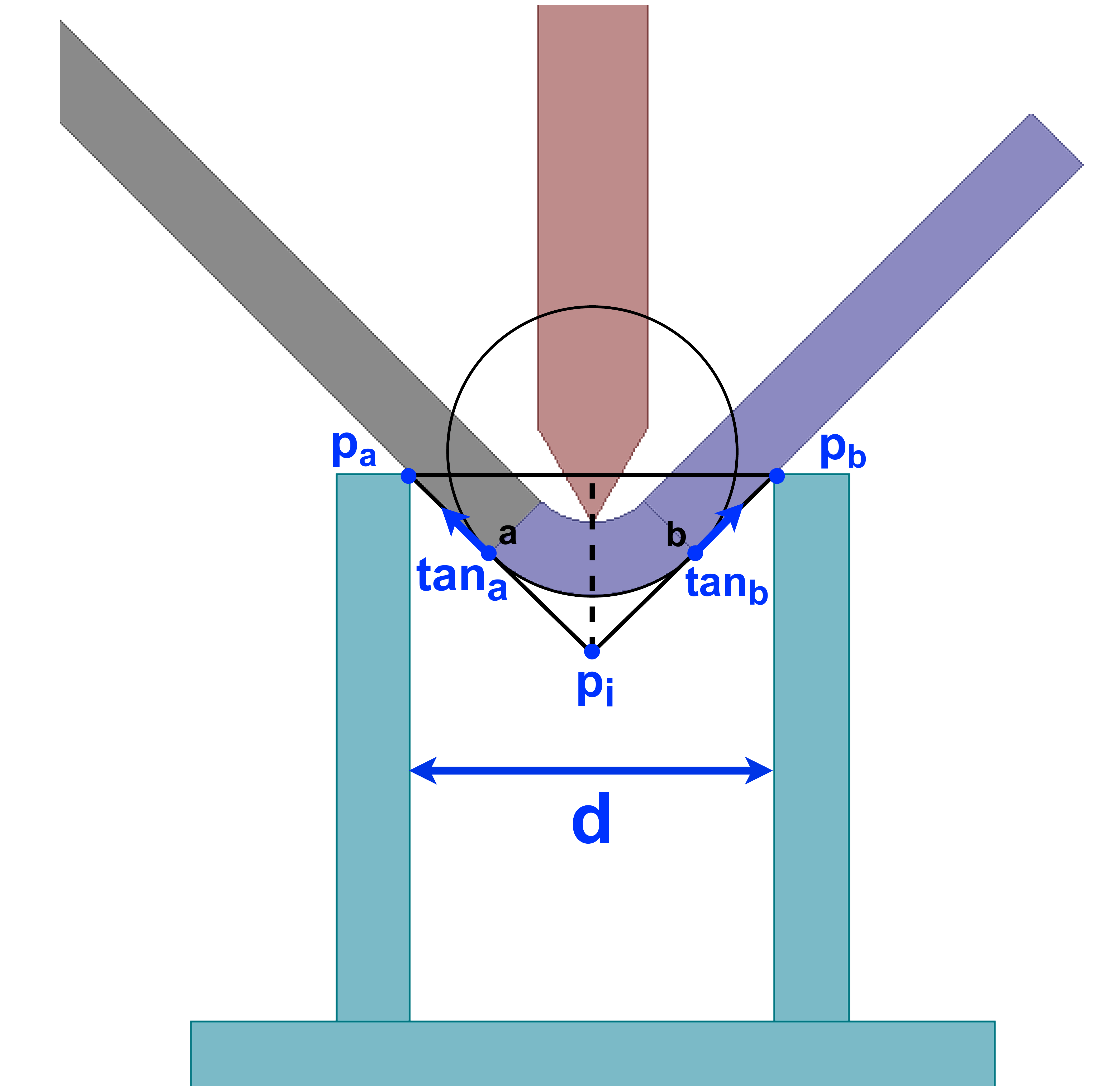}
		\caption{Die construction}\label{fig:die_creation}
	\end{subfigure}
	\caption{Punch and die geometries are built around the bend center using bend parameters and simple trigonometry.}\label{fig:tooling_creation}
\end{figure}

While the standard bend simulation procedure produces geometrically diverse parts, it does not inherently enforce manufacturability constraints that would prevent certain designs from being feasible in practice. 
To emulate realistic, configuration-dependent manufacturability metrics, BenDFM explicitly models the interaction between part geometry and bending tooling.
In sheet metal bending, a punch (the upper tool) presses material downward while a die (the lower tool with a channel) provides backing support and defines the bend geometry. 
To mimic this interaction, both punch and die geometries are parametrically created around each bend during generation, as illustrated in Figure~\ref{fig:punch_and_die}. 
Tooling is instantiated at the midpoint of each bend line, with dimensions and positions determined by both local bend parameters and predefined tooling specifications. 
This midpoint-centric approach ensures consistent and accurate alignment between the tooling and the bend geometry, regardless of individual bend parameters.

The punch creation process, shown in Figure~\ref{fig:punch_creation}, uses the punch tip angle and punch thickness in conjunction with bend parameters to define the tool geometry. 
Starting from the midpoint of the inner bend arc ($p_m$), two direction vectors $\vec{a}$ and $\vec{b}$ are computed based on the punch tip angle. 
These vectors form an isosceles triangle with its apex at $p_m$. 
The base of this triangle connects points $p_a$ and $p_b$ along the vectors $\vec{a}$ and $\vec{b}$. 
Given the punch thickness $t$ and the punch tip angle, the positions of $p_a$ and $p_b$ are determined using basic trigonometry.
The triangle is then extruded along the bend edge direction to form the punch tip, and additional rectangular blocks are extruded upward to complete the punch body.

The die creation follows a similar procedure, as illustrated in Figure~\ref{fig:die_creation}. 
The key parameter here is the die opening width $d$, which determines the lateral spacing between the outer surfaces of the bent part and the die walls. 
During bending, the part is pushed into the middle of the die channel, which means the die is always in contact with the outer contour of the part with equal spacing on both sides.
To ensure this, tangents are drawn at the endpoints of the bend arc to identify their intersection point ($p_i$), from which two points $p_a$ and $p_b$ are located at a distance of $d/2$ from the bend centerline, forming an isosceles triangle. 
Trigonometric relationships are again used to compute the positions of $p_a$ and $p_b$, which then define the die channel geometry. 
As with the punch, extrusion operations are used to form the rectangular blocks that make up the base of the die.

\subsection{Dynamic bend modeling and unfolding}\label{sec:intermediate_bends}

Adding bends via discrete extrusions at the final angle, rather than simulating continuous deformation, means that tooling geometries would only be evaluated at the bend's final state. 
In reality, however, the bending process is dynamic: the punch and die move relative to the sheet along the bend line, progressively deforming the material.
To accurately reflect this behavior, BenDFM emulates the dynamic nature of the bend path by generating each new flange (and its corresponding punch and die geometries) at discrete angular intervals between $0^\circ$ and the final bend angle. 
This allows the evolution of the part geometry and tooling interaction to be captured more realistically at intermediate states.

Representing these intermediate configurations requires material deformation to be accounted for using bend allowances. 
During the bending process, the outer surface of the material stretches while the inner surface compresses. 
As a result, the flat length needed to form a given flange is not simply equal to the sum of its internal or external dimensions. 
A bend allowance (BA) is a compensation factor that accounts for this material deformation, ensuring that the flat pattern (the 2D layout from which material is cut before bending) has the correct dimensions. 
The bend allowance is computed as:

\[
\mathrm{BA} = \frac{\pi}{180} \cdot \theta \cdot \left( r + K \cdot t \right)
\]
where $\theta$ is the bend angle (in degrees), $r$ is the inside bend radius, $t$ is the sheet thickness, and $K$ is a material-specific factor typically ranging from 0.3 to 0.5.

\begin{figure}[htp]
	\centering
	\begin{subfigure}[b]{0.38\textwidth}
		\centering
		\includegraphics[width=\textwidth]{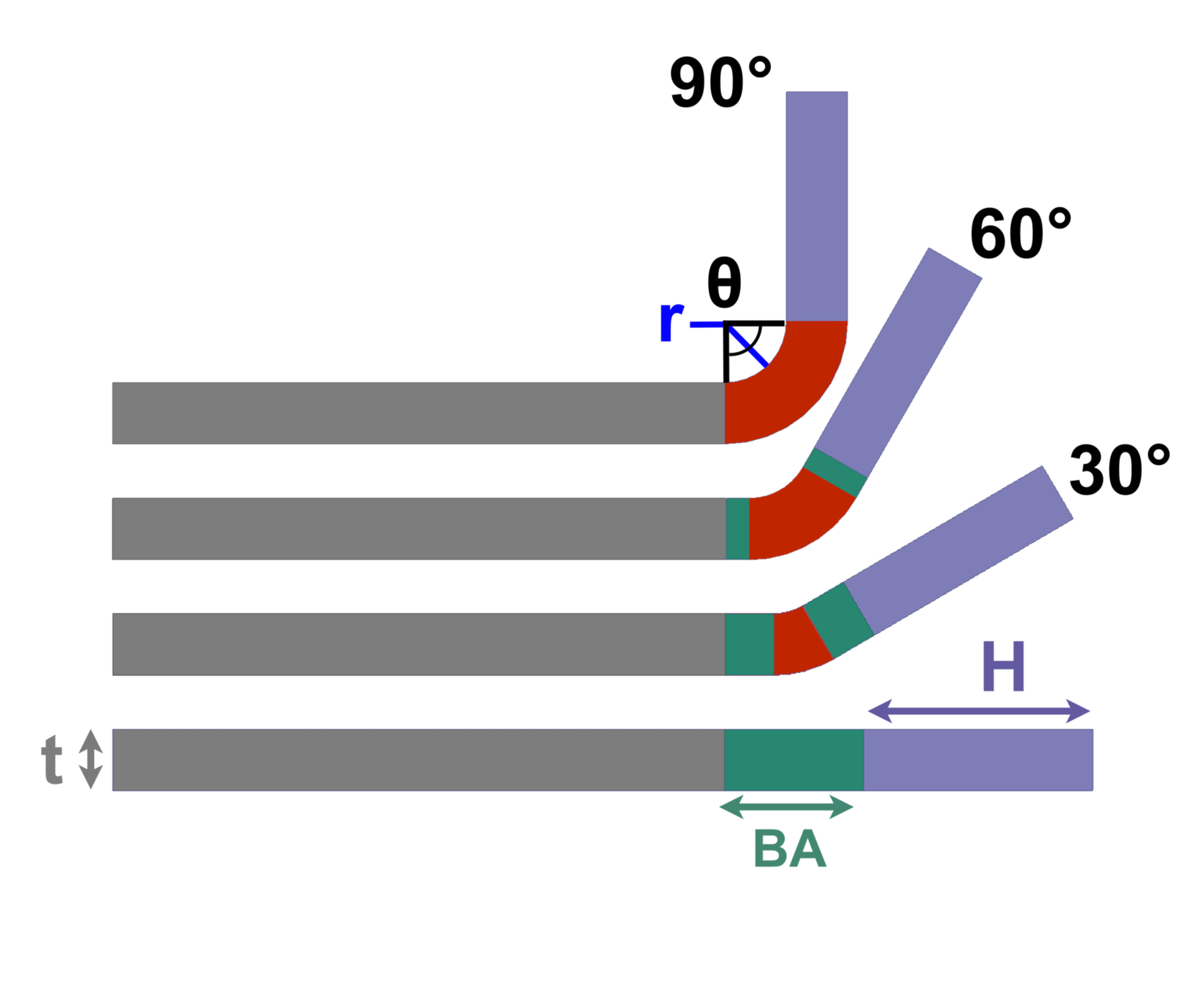}
		\caption{Bend allowance at intermediate states.}\label{fig:allowance}
	\end{subfigure}
	\hspace{0.03\textwidth}
	\begin{subfigure}[b]{0.57\textwidth}
		\centering
		\includegraphics[width=\textwidth]{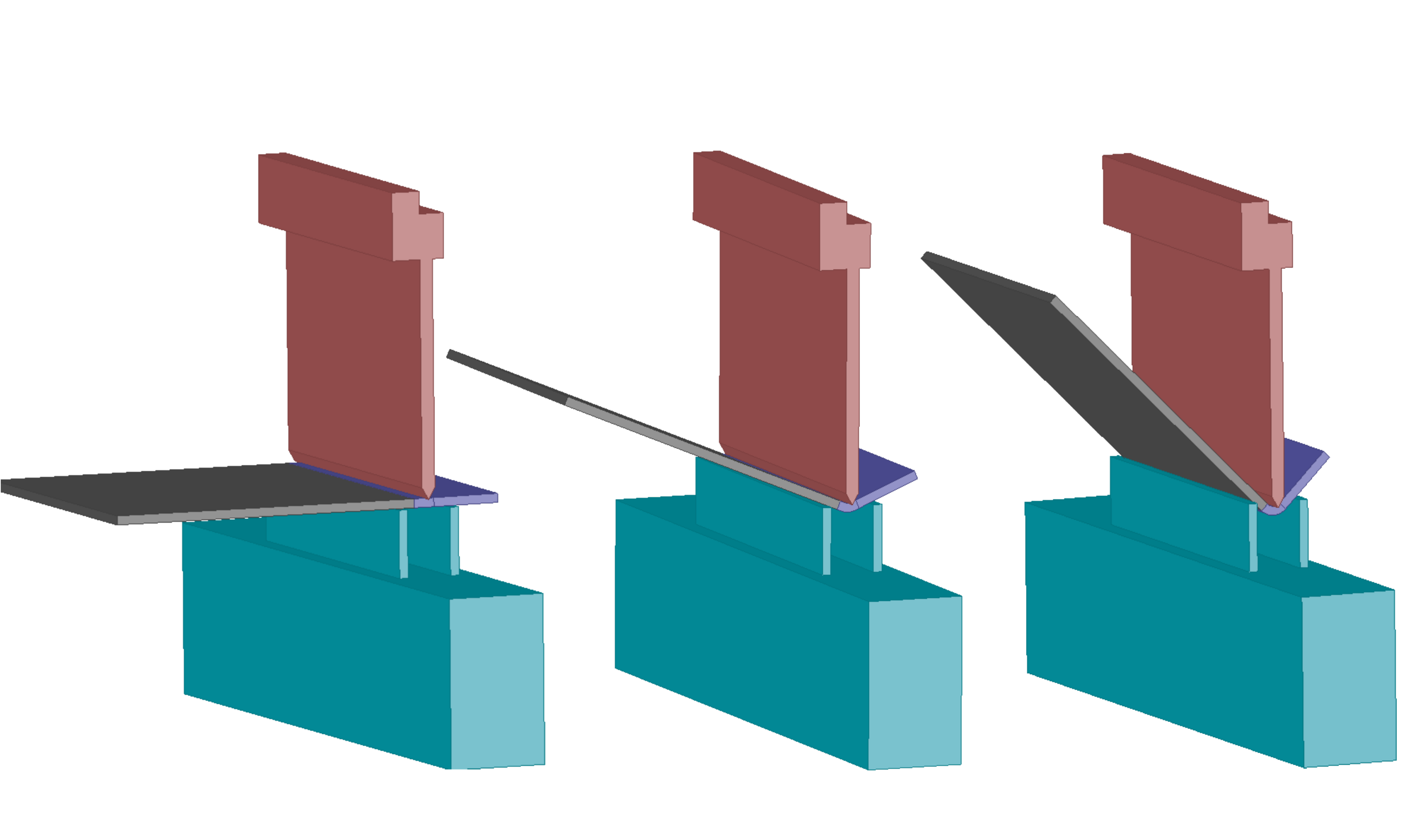}
		\caption{Tooling at intermediate states ($0^\circ$, $45^\circ$, and $90^\circ$).}\label{fig:path_2}
	\end{subfigure}
		\caption{(a) As the bend progresses, a greater portion of the bend allowance is incorporated into the bend itself (red), while the remainder is distributed as flat flanges (green) on either side. 
		(b) Punch and die tools are dynamically positioned relative to the evolving part geometry throughout the bend trajectory, shown here for $0^\circ$, $45^\circ$, and $90^\circ$ angles.}\label{fig:intermediate_bends}
	\end{figure}

To incorporate the bend allowance during generation, once the total bend allowance $\text{BA}(\theta)$ is computed for the target angle, the geometry at any intermediate angle $\theta_i$ is created by proportionally distributing the allowance. 
A portion of the total allowance, $\text{BA}(\theta_i)$, is realized in the partially formed bend segment (shown in red in Figure~\ref{fig:allowance}). 
The remaining allowance, $\text{BA}(\theta) - \text{BA}(\theta_i)$, is distributed equally as flat flanges on either side of the bend (green in Figure~\ref{fig:allowance}). 
This ensures that, at each step, the overall bend allowance is preserved and the bend remains centered within it. 
At $\theta_i = 0^\circ$, the entire allowance resides in the flat flanges; at $\theta_i = \theta$, the full allowance is incorporated in the bend itself. 
This enables the accurate dynamic positioning of tooling geometries throughout the bend path, as illustrated in Figure~\ref{fig:path_2}.

\begin{figure}[htp]
	\centering
	\includegraphics[width=0.45\textwidth]{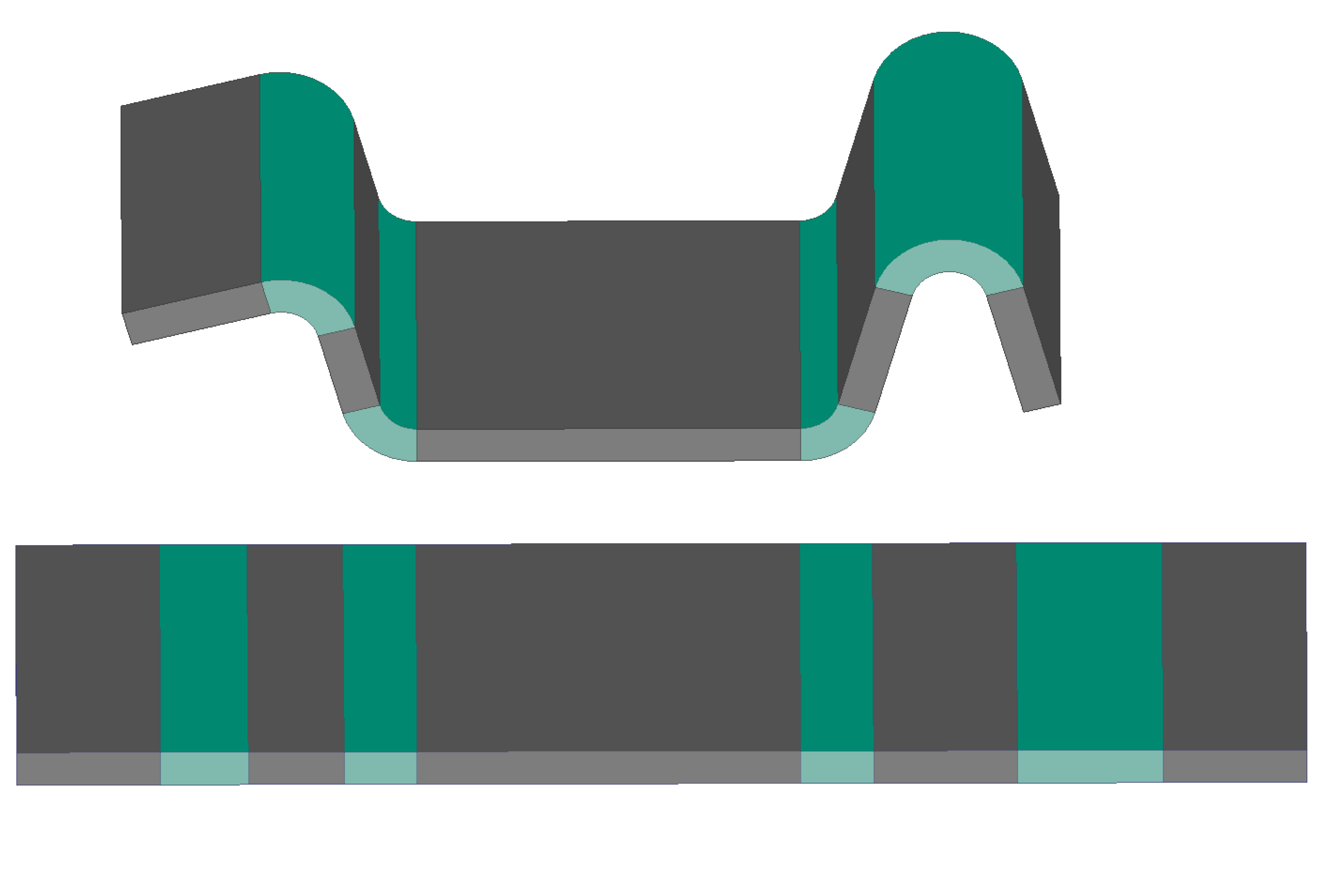}
	\caption{Example of a bent part and its corresponding unfolded flat pattern, accounting for material deformation during bending using bend allowances.}\label{fig:unfolding}
	\end{figure}

Bend allowances are crucial in practice to accurately represent the part's unfolded geometry. 
Unfolding is a critical operation in sheet metal fabrication that transforms a 3D bent part into its 2D flat pattern. 
This flat pattern represents the precise 2D layout from which material is cut prior to bending, accounting for all material deformation that occurs during the bending process. 
In BenDFM, unfolding is performed by replaying the full bend sequence with all angles set to $0^\circ$, generating both the bend-allowance flanges from the original angles and the flat flanges for each bend. 
The resulting unfolded geometry, as shown in Figure~\ref{fig:unfolding}, is stored alongside the 3D model and bend sequence metadata, providing a complementary 2D representation of the part.
This unfolded view is essential for determining material requirements and is also critical for related tasks such as sheet metal nesting (layout optimization) and cutting.

\subsection{Manufacturability label generation}
\subsubsection{Feasibility labels}

\begin{figure}[ht]
	\centering
	\begin{subfigure}[b]{0.35\textwidth}
		\centering
		\includegraphics[width=\textwidth]{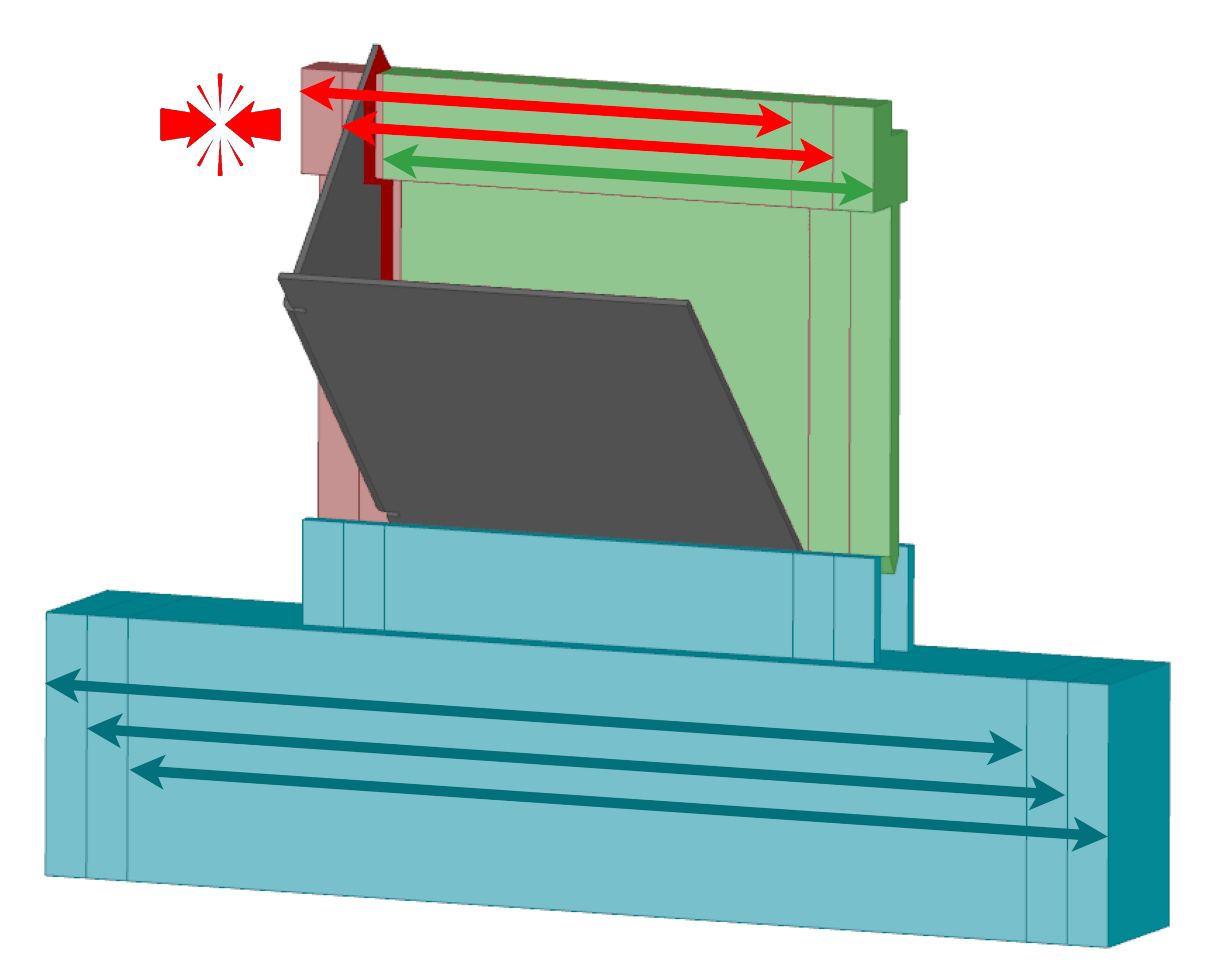}
		\caption{Tool alignment flexibility}\label{fig:tool_alignment}
	\end{subfigure}
	\hspace{0.1\textwidth}
	\begin{subfigure}[b]{0.35\textwidth}
		\centering
		\includegraphics[width=\textwidth]{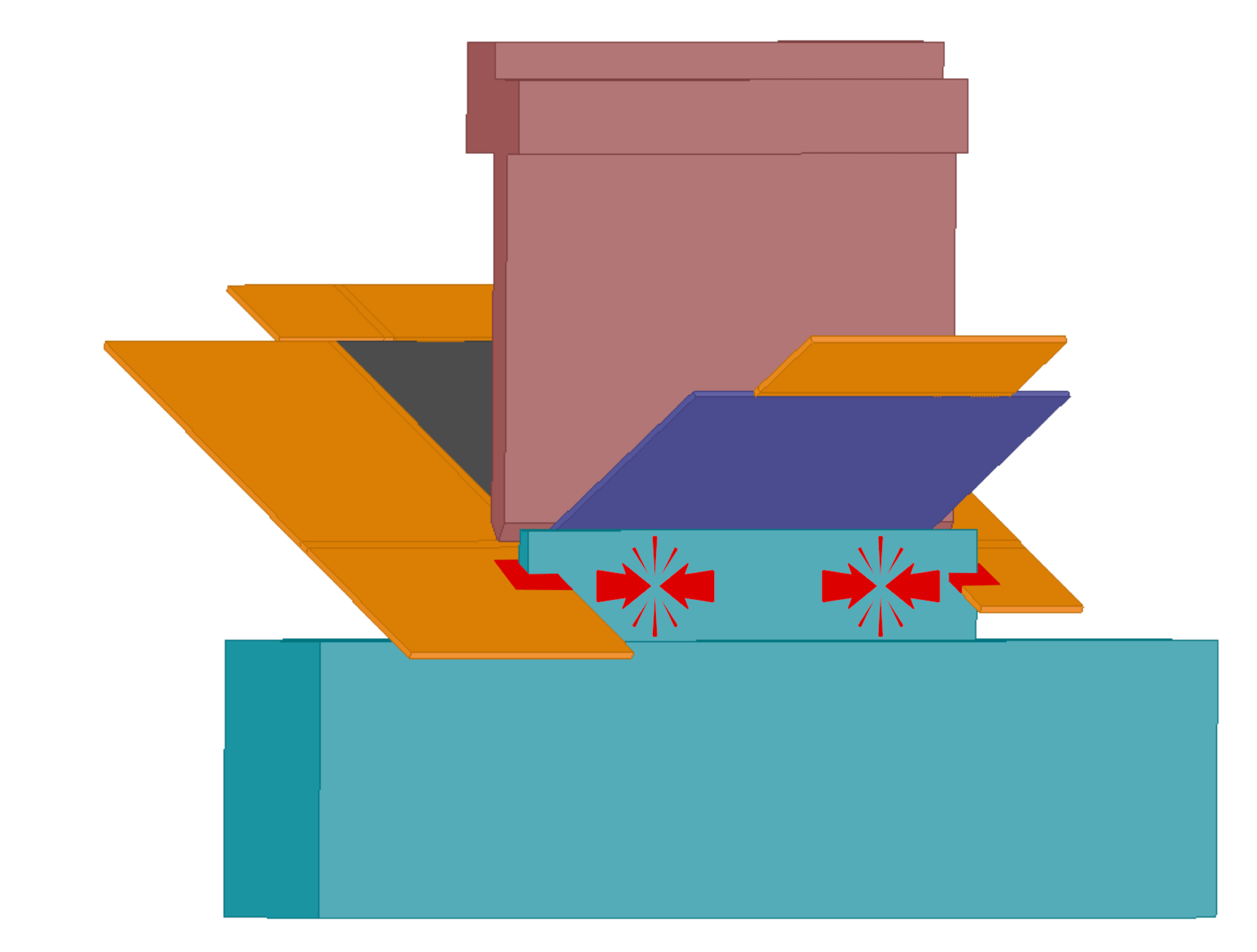}
		\caption{Post-hoc collision check}\label{fig:post_hoc_collision_check}
	\end{subfigure}
	\caption{(a) Placing the punch on the left or middle of the bend (depicted in red) results in a collision, offsetting it to the right (depicted in green) avoids this. 
	(b) Adding future bends in their flat state (depicted in orange), reveals a die collision with later flanges during the first bend (in purple).}\label{fig:extra_features}
\end{figure}

BenDFM captures two configuration-dependent feasibility labels: punch-part collisions and die-part collisions. 
Using the methodology described above, intermediate flanges and corresponding punch and die geometries are instantiated in $5^\circ$ increments along each bend's trajectory to allow for collision checks throughout the bending motion.
Collision detection is performed by computing intersection volumes between tooling and the part; any overlap is registered as a collision.
To ensure collision accuracy, two enhancements are introduced in the bending procedure, as illustrated in Figure~\ref{fig:extra_features}.

First, punch and die geometries are instantiated at three positions along the bend line (left, center, and right) to reflect real-world alignment flexibility (Figure~\ref{fig:tool_alignment}). 
In practice, a part is not always centered on the tooling, but can be shifted along the punch axis. 
By incorporating this positional flexibility, our method avoids false positive collisions that could be prevented by slightly repositioning the part.
A collision is only registered if none of the three alignment options are collision-free along all of the intermediate bend angles.

Second, collision detection is performed after the complete part geometry has been generated, as shown in Figure~\ref{fig:post_hoc_collision_check}.
A post-hoc approach is necessary because flanges are added sequentially: a bend that appears collision-free when created (depicted in purple) may later become infeasible due to interference with subsequently added flanges (depicted in orange). 
To capture these interactions, collision checks retrace each bend and evaluate collisions across all intermediate bend angles (as discussed in Section~\ref{sec:intermediate_bends}) while incorporating all unbent downstream flanges and their bend allowances.
This procedure ensures that the final collision labels accurately reflect the global geometry after all bends are added.

\begin{figure}[htp]
	\centering
	\includegraphics[width=0.7\textwidth]{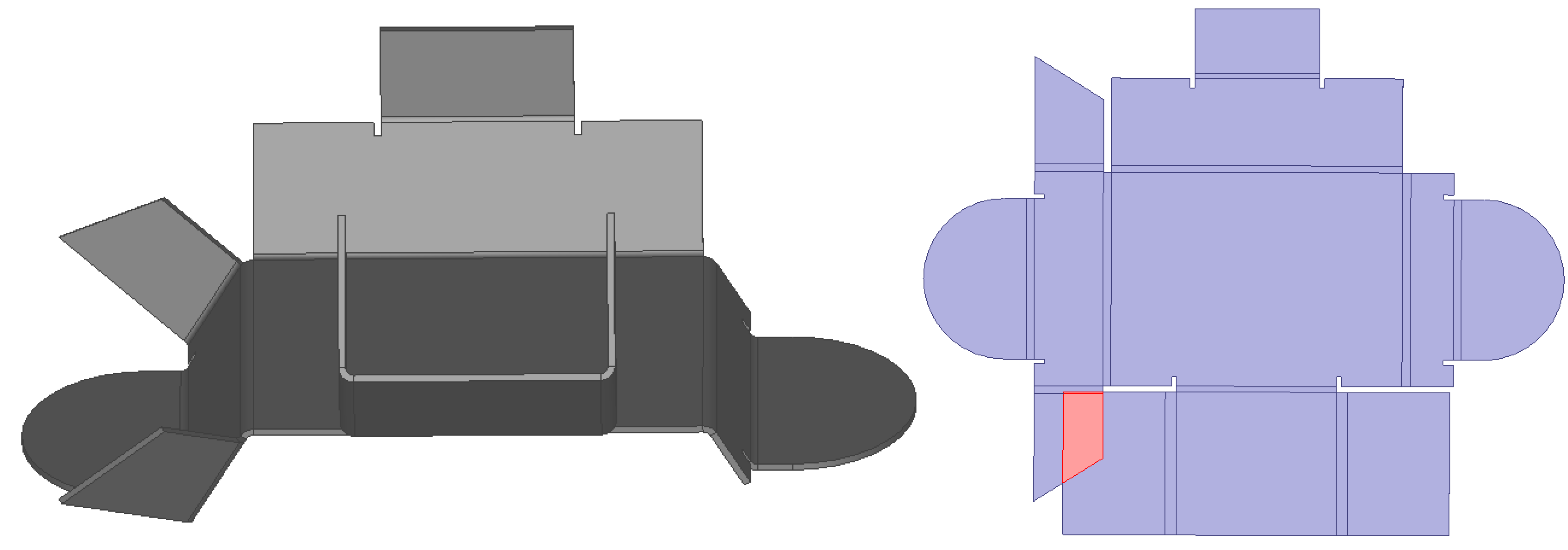}
	\caption{Example of a part whose unfolded flat pattern exhibits self-intersection, making it impossible to cut from a single sheet.}\label{fig:unfolding_self_intersection}
\end{figure}

In addition to configuration-dependent feasibility labels, BenDFM includes a configuration-independent feasibility check for unfolding overlaps (Figure~\ref{fig:unfolding_self_intersection}). 
These occur when the unfolded flat pattern of a part intersects with itself, making it impossible to cut the design from a single flat sheet regardless of tooling configuration. 
This check is performed by unfolding the part (Section~\ref{sec:intermediate_bends}) and detecting overlapping volumes using geometric B-rep boolean operations.

\subsubsection{Complexity metrics} \label{sec:complexity_metrics}

\begin{figure}[htp]
	\centering
	\includegraphics[width=0.4\textwidth]{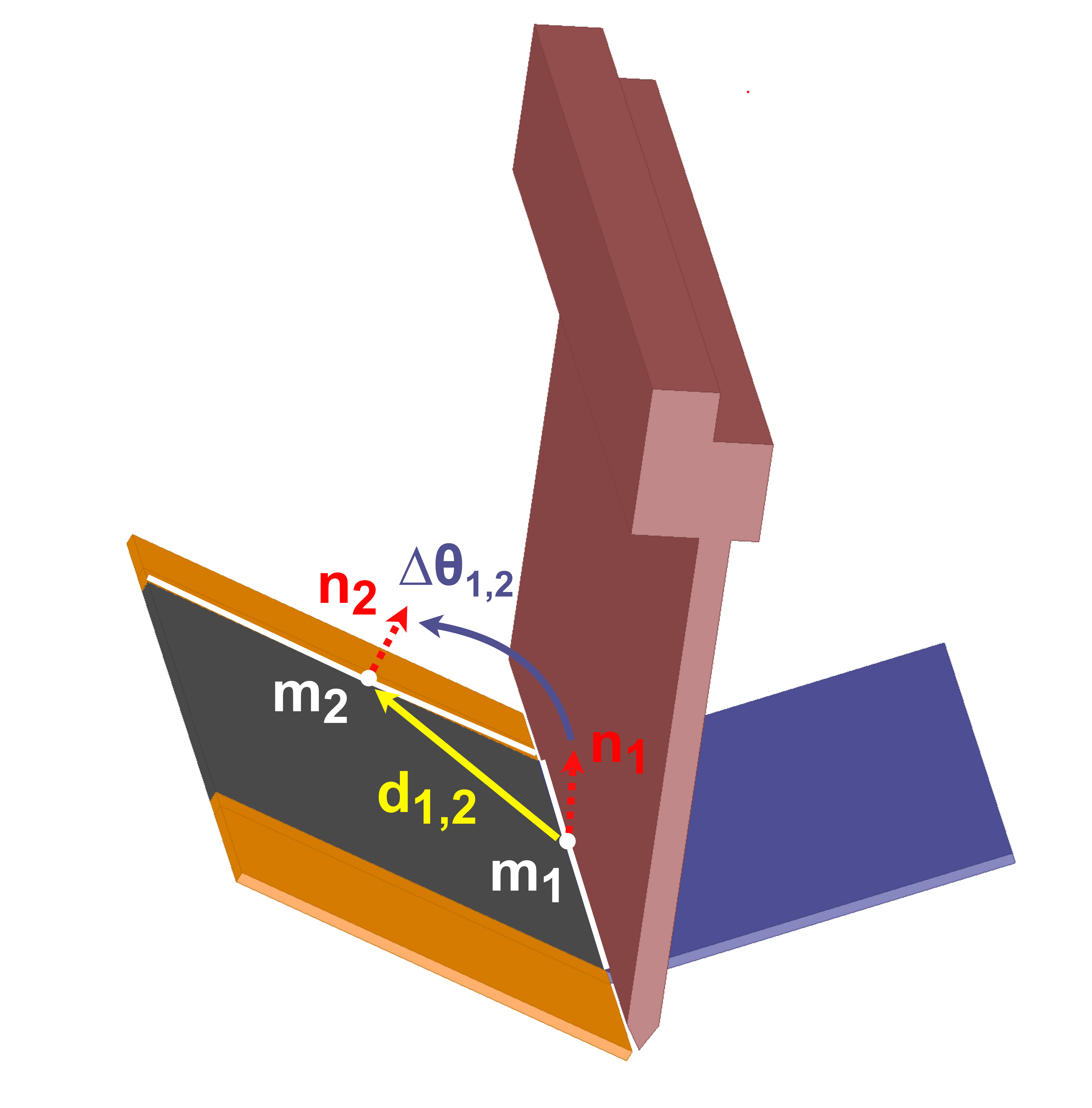}
	\caption{Illustration of part reorientation metrics: the travel distance between consecutive bends (yellow) and the rotation angle required to align the part for the next bending operation (purple).}
	\label{fig:reorientation}
\end{figure}

Beyond feasibility labels, BenDFM provides a set of metrics that quantify both configuration-dependent and configuration-independent complexity of the generated parts.
Configuration-dependent complexity arises from the need to reorient the part between successive bends, as each tooling operation must align along the corresponding bend line. 
This reorientation can be decomposed into a translational shift and a rotational movement. 
Accordingly, we introduce three metrics: part reorientation distance, reorientation angle, and a derived part flip indicator, as illustrated in Figure~\ref{fig:reorientation}. 
These metrics quantify the handling effort required during manufacturing, which is particularly relevant for large or heavy parts.
These metrics are computed between consecutive bends, with their sum over the entire bend sequence yielding aggregated complexity scores for the full part. 
Since these metrics depend on the specific bend sequence rather than the geometry alone, they are inherently configuration-dependent.

For each bend $i$, let $\mathbf{m}_i$ denote the midpoint of the bend edge. 
The reorientation distance $d_{i,i-1}$ between bends $i$ and $i-1$ is defined as the Euclidean distance between their respective midpoints:
\[
d_{i,i-1} = \| \mathbf{m}_i - \mathbf{m}_{i-1} \|
\]

Let $\mathbf{n}_i^\theta$ be the normal vector from the center of the bend arc at angle $\theta$, indicating the punch direction during bending. 
The reorientation angle $\Delta\theta_{i,i-1}$ is then defined as the angle between the punch direction at the end of bend $i-1$ ($\mathbf{n}_{i-1}^\theta$) and the starting punch direction for bend $i$ ($\mathbf{n}_i^0$):
\[
\Delta\theta_{i,i-1} = \arccos\left( \mathbf{n}_{i-1}^\theta \cdot \mathbf{n}_i^0 \right)
\]

To capture more practical implications, we record a \textit{part flip} when the reorientation angle exceeds $90^\circ$. 
While actual flipping may also depend on other geometric constraints not captured by this measure, this metric offers an interpretable approximation. 

In terms of configuration-independent complexity metrics, the generation process allows us to record a wide variety of features.
At the part level, we record features including: the number of bends; sheet thickness; 3D bounding-box volume ($\mathrm{cm}^3$); solid part volume ($\mathrm{cm}^3$); unfolded rectangular bounding area ($\mathrm{cm}^2$); and part mass (kg). 
Part mass is computed from the solid volume using the material density of steel ($\rho = 7850\ \mathrm{kg}/\mathrm{m}^3$), but can be easily altered using the logged solid volume. 
From the individual per-bend parameters, other aggregate statistics can easily be computed, including but not limited to: the number of distinct bend angles and radii, the minimum and maximum flange height, bend angle, and radius, the number of bends with bend reliefs, and the number of bends with rounded flanges.

\subsection{Dataset overview}
The full BenDFM dataset comprises 20,000 unique 3D bent part geometries in STEP format. 
Each part is accompanied by its corresponding unfolded model in STEP format, a complete bend sequence with per-bend parameters, and part-level manufacturability labels and metadata in a JSON file. 
The generation parameters were chosen to reflect general industrial practices. 
Base sheet dimensions are uniformly sampled, with length and width ranging from 150 mm to 300 mm, and thickness from 2.0 mm to 6.0 mm. 
Bend angles are drawn from $\{45^\circ, 60^\circ, 90^\circ, 120^\circ, 135^\circ\}$, with a bias toward the very common $90^\circ$ angle. 
Flange heights are sampled between 75 mm and the maximum base sheet dimension. 
To enhance realism, 20\% of bends are created with partial widths (50-75\% of edge length), which triggers the automatic insertion of bend reliefs. 
Bend radii are sampled proportionally to sheet thickness ($1.0$--$1.5\times$), and each bend is randomly oriented upward or downward.
All parts are evaluated using a fixed punch and die geometry, with a punch thickness of 10 mm, a punch tip angle of $90^\circ$, a length of 300 mm, and a die opening width of 40 mm.  

While feasibility is often simplified to a binary classification task in the literature~\citep{ghadai2018learning}, our taxonomy recognizes it can also be quantified by severity (e.g., the number of infeasible bends). 
Although our benchmark experiments adopt the conventional binary approach at the part level, the BenDFM dataset is designed to support both perspectives. 
It provides granular per-bend collision flags and aggregated collision counts alongside binary part-level labels, enabling a full spectrum of tasks from binary classification to severity prediction.
To ensure that the severity of tooling collision violations is controlled, 40\% of bends are guaranteed to be collision-free during generation by resampling their parameters until a feasible bend is found. 

This prevents the dataset from being populated with trivially unmanufacturable designs with an overwhelming number of colliding bends and instead establishes a more subtle and challenging decision boundary for learning models.
The full dataset, detailed characteristics, and sample part visualizations will be made publicly available on GitHub\footnote{\url{https://github.com/UGent-CVAMO/bendfm}}.

To facilitate benchmarking, the 20,000-part dataset is divided into two non-overlapping subsets, each balanced for a specific feasibility classification task. 
This partitioning is necessary because the two failure modes under consideration—tooling collisions and unfolding overlaps—occur at different natural rates during generation. Creating distinct, balanced subsets for each task facilitates model evaluation.

The primary subset, \textbf{BenDFM}, is designed for benchmarking the prediction of \textit{tooling collisions}. 
It contains 14,000 parts with bend counts ranging from 2 to 8. 
The subset is balanced with a 50/50 split between parts with and without tooling collisions. 
This balancing is performed via stratified sampling for each bend count (e.g., 1,000 collision-free and 1,000 colliding parts for the 4-bend group), which prevents models from learning a spurious correlation between the number of bends and the likelihood of collision. 
To isolate the configurational feasibility task, all parts in this subset are guaranteed to be free of unfolding overlaps.

The second subset, \textbf{BenDFM-U}, is tailored for the \textit{unfolding overlap} task.
Since overlaps in the flat pattern typically arise in more complex parts, this subset consists of 6,000 parts with higher bend counts (7 to 10). 
It is similarly balanced with a 50/50 split between parts with and without unfolding overlaps, stratified by bend count. 
This focused subset provides a clear and balanced benchmark for evaluating a model's ability to implicitly learn the unfolding transformation and detect geometric impossibilities.

\section{Benchmarking results on BenDFM}\label{sec:benchmark}
The BenDFM dataset presents novel challenges for deep learning-based manufacturability assessment, owing to the often structural, global geometric factors that influence bending feasibility and complexity.
To establish baseline performances on this new public resource, we benchmark two state-of-the-art geometric deep learning models on a representative metric from each of the four taxonomy quadrants.
Feasibility labels are treated as classification tasks: we predict the presence of tooling collisions (configurational feasibility) on the BenDFM subset, and unfolding overlaps (geometric feasibility) on the BenDFM-U subset.
The former requires the model to learn complex interactions between geometry and tooling configurations, while the latter demands an implicit understanding of the unfolding transformation from 3D to 2D.

Complexity labels are continuous and evaluated via regression tasks: we predict the number of part flips (configurational complexity) and the unfolded part's required rectangular sheet area (geometric complexity) on the BenDFM subset.
The former serves as a proxy for handling effort, while the latter is a key determinant of material utilization that again requires the model to implicitly learn the unfolding operation.

\subsection{Input representations and model architectures}

The designs in BenDFM are provided as STEP files, which use boundary representations (B-reps) to define 3D geometry. 
B-reps represent solid geometry through topological elements—faces (surfaces), edges, and vertices—rather than as solid volumetric data, making them the standard representation format in CAD systems. 
Since B-reps are not directly compatible with standard neural network architectures, they must be converted into suitable input formats. 
We benchmark two of the most prominent representations used in geometric deep learning for manufacturing applications: point clouds and graphs, pairing each with a state-of-the-art model architecture.

Point clouds are unordered sets of 3D points sampled from a part's surface, representing geometry without explicit connectivity information. 
This representation is popular for 3D deep learning, with demonstrated success in manufacturing applications like feature recognition~\citep{zhang2022machining, zhang2024point} and process selection~\citep{wang2023manufacturing2,liu2025manufacturing}. 
We benchmark PointNext~\citep{qian2022pointnext}, a state-of-the-art architecture from the PointNet family.
PointNext processes point clouds hierarchically, learning features from local neighborhoods at progressively larger scales. 
It improves upon prior PointNet models through more efficient neural network blocks and improved training strategies.
A key property is that it remains invariant to the order of input points, making it robust to different point orderings.
Following the original implementation, we sample 1024 points per part, using their 3D coordinates and surface normals as input features. 
 
Graphs are another powerful representation for 3D geometry, particularly well-suited for capturing relational structures.
B-reps can be represented as attributed adjacency graphs (AAGs), where each face becomes a node in a graph, adjacencies between faces become edges, and geometric properties (such as surface curvature or area) are encoded as node and edge attributes. 
This representation preserves both the topological structure and geometric information of the original B-rep without the information loss associated with point cloud discretization. 
We benchmark UV-Net~\citep{jayaraman2021uv}, a foundational architecture for B-rep learning that has inspired many subsequent models in manufacturing applications~\citep{li2025classification, hussong2025selection, zheng2025sfrgnn, niu2024manufacturing, zhong2025deepmill}.
UV-Net operates by sampling points on each face and edge within their native parametric domains. 
Specifically, each face is sampled into a 2D grid of points in its UV space (the 2D parameter space of the surface), and each edge into a 1D sequence of points in its U space (the 1D parameter space of the curve). 
For each sampled point, a feature vector is constructed containing its 3D coordinates, surface normal, and trim flag.
These grids of features are then processed by 2D and 1D Convolutional Neural Networks (CNNs) to produce fixed-size embeddings for each face and edge. 
Finally, these embeddings are used as node and edge attributes in a Graph Convolutional Network (GCN) that operates on the part's AAG to learn a global part representation.
Following the original implementation~\citep{jayaraman2021uv}, we use a $10 \times 10$ grid for faces and a 10-point sequence for edges.

\subsection{Experimental setup, training, and evaluation}

We split the dataset into training (80\%), validation (10\%), and test (10\%) sets, using a fixed split to ensure reproducibility across models. 
All models are trained using the Adam optimizer with a learning rate of $0.0005$, a batch size of 32, and early stopping based on validation performance with a patience of 20 epochs.
To ensure robust evaluation, each experiment is repeated five times with fixed random seeds for weight initialization and data shuffling, following best practices for reproducible deep learning~\citep{henderson2018deep}. 
The same seeds are used across both models to ensure equal data splits and initialization conditions for fair comparison.

Classification models are trained to minimize binary cross-entropy loss:

\[
\mathcal{L}_{\text{BCE}} = -\frac{1}{N} \sum_{i=1}^N \left[ y_i \log(\hat{y}_i) + (1 - y_i) \log(1 - \hat{y}_i) \right]
\]

where $y_i \in \{0,1\}$ denotes the ground-truth label of sample $i$ and $\hat{y}_i \in [0,1]$ represents its predicted probability.
Classification performance is evaluated using three complementary metrics. 
First, the Area Under the ROC Curve (AUC) assesses the model's overall threshold-independent discriminative power. 
Second, accuracy offers a straightforward measure of correctness on the (balanced) dataset. 
Finally, the F1-score, the harmonic mean of precision and recall, provides a balanced assessment of a model's ability to correctly identify positive cases while minimizing false alarms.
All three metrics are bounded between 0 and 1, with higher values indicating better performance.

For regression tasks, models are trained using mean squared error (MSE) loss:
\[
\mathcal{L}_{\text{MSE}} = \frac{1}{N} \sum_{i=1}^N {(y_i - \hat{y}_i)}^2
\]

Regression performance is evaluated using three complementary metrics. 
Mean absolute error (MAE) provides a direct, interpretable measure of the average prediction error. 
Root mean squared error (RMSE) is also in the original units but penalizes larger errors more heavily. 
Finally, mean absolute percentage error (MAPE) offers a scale-independent view of relative accuracy.
All three metrics are non-negative, with lower values indicating better performance.

\subsection{Binary classification of feasibility labels}

To assess the models' ability to capture different types of manufacturability constraints, we formulate two binary classification tasks, each corresponding to one of the feasibility quadrants in our proposed taxonomy.
The first task evaluates the prediction of tooling collisions, an instance of \textit{configurational feasibility}, on the BenDFM subset. 
A part is labeled as infeasible (positive class) if any bend in its sequence results in a collision.
The second task assesses the detection of unfolding overlaps, an instance of \textit{geometric feasibility}, on the BenDFM-U subset. 
A part is labeled as infeasible if its unfolded flat pattern self-intersects, requiring the model to implicitly learn the transformation from the 3D part to its 2D unfolded state from the 3D geometry alone.
Given the balanced nature of the test sets, a standard classification threshold of 0.5 is used to derive accuracy and F1-scores.
The results for UV-Net and PointNext are presented in Table~\ref{tab:collision_results}, alongside a null model baseline that randomly guesses the class with equal probability.

\begin{table}[htp]
	\centering
	\small
	\renewcommand{\arraystretch}{1.4}
	\caption{Feasibility classification results. 
	Tool collision detection is evaluated on the BenDFM subset; unfolding overlap detection is evaluated on the BenDFM-U subset.
	Reported values are means (standard deviations) over 5 runs with different random seeds.}\label{tab:collision_results}
	\begin{tabular}{lcccccc}
		\toprule
		& \multicolumn{3}{c}{\textbf{Tool collision}} & \multicolumn{3}{c}{\textbf{Unfolding overlap}} \\
		\cmidrule(lr){2-4} \cmidrule(lr){5-7} 
		\textbf{Model} & \textbf{AUC} & \textbf{Acc (\%)} & \textbf{F1 (\%) }& \textbf{AUC} & \textbf{Acc (\%)} &\textbf{ F1 (\%)} \\
		\midrule
		UV-Net     & 0.840 (0.003)  & 76.07 (0.43) & 75.41 (0.06) & 0.896 (0.009) & 81.80 (1.37) & 81.31 (1.93) \\
		PointNext  & 0.827 (0.003)  & 73.83 (0.10) & 71.74 (1.27) & 0.844 (0.008) & 76.13 (1.21) & 76.55 (0.53) \\
		\hdashline
		Baseline & 0.500 & 50.00 & 50.00 & 0.500 & 50.00 & 50.00 \\
		\bottomrule
	\end{tabular}
\end{table}

Both UV-Net and PointNext demonstrate significant improvements over the null model baseline across all feasibility prediction tasks.
For the configurational feasibility task (tooling collision detection), UV-Net achieves superior performance with an AUC of 0.840 and 76.07\% accuracy, compared to PointNext's AUC of 0.827 and 73.83\% accuracy.
This performance advantage is more pronounced for the geometric feasibility task (unfolding overlap detection), where UV-Net reaches an AUC of 0.896 and 81.80\% accuracy versus PointNext's AUC of 0.844 and 76.13\% accuracy.
Notably, both models exhibit stronger performance on geometric feasibility compared to configurational feasibility.

\subsection{Regression of complexity labels}
To evaluate model performance on the complexity quadrants of our taxonomy, we formulate two regression tasks using the BenDFM subset.
The first task predicts the number of part flips (as defined in Section~\ref{sec:complexity_metrics}), representing \textit{configurational complexity}. 
This integer-valued metric quantifies the handling effort required during manufacturing, with model predictions rounded to the nearest integer for evaluation.
For parts containing 2--8 bends, the target values range from 0 to 7 flips.
The second task predicts the unfolded part's required sheet area (in $cm^2$), a measure of \textit{geometric complexity}. 
We compute this as the area of the 2D rectangular bounding box of the unfolded flat pattern (including bend allowances). 
This configuration-independent metric quantifies material usage and, like for the unfolding overlap task, requires the model to implicitly learn the geometric transformation from bent geometry to unfolded representation.

For both regression tasks, we establish a naive baseline using a constant predictor that outputs the training set mean for all test samples (1.37 for part flips; 254.59 $cm^2$ for unfolded bounding area).
This baseline provides a lower bound for meaningful model performance.
To account for the different measurement scales between tasks, we report both absolute error metrics and relative percentage improvements over the baseline.
The coefficient of variation for part flips (0.469) and unfolded bounding area (0.436) indicates that both tasks exhibit similar inherent variability in the dataset.
Results for both tasks are presented in Table~\ref{tab:regression_results}.

\begin{table}[htp]
	\renewcommand{\arraystretch}{1.4}
	\small
	\centering
	\caption{Complexity regression results. 
	Both tasks are evaluated on the BenDFM subset. 
	Reported values are means (standard deviations) over 5 runs with different random seeds.}\label{tab:regression_results}
	\begin{tabular}{lcccccc}
		\toprule
		& \multicolumn{3}{c}{\textbf{Part flips}} & \multicolumn{3}{c}{\textbf{Unfolded bounding area}} \\
		\cmidrule(lr){2-4} \cmidrule(lr){5-7}
		\textbf{Method} & \textbf{MAE} & \textbf{RMSE} & \textbf{MAPE (\%)} & \textbf{MAE} & \textbf{RMSE} & \textbf{MAPE (\%)} \\
		\midrule
		UV-Net  & 0.54 (0.01) & 0.86 (0.02) & 35.52 (0.90) & 14.60 (0.86) & 20.68 (1.17) & 5.90 (0.41) \\
		PointNext  & 0.59 (0.01) & 0.90 (0.01) & 39.33 (1.70) & 20.24 (0.93) & 28.82 (1.37) & 8.28 (0.49) \\
		\hdashline
		Baseline & 0.984 & 1.180 & 67.67 & 89.81 & 111.35 & 46.01 \\
		\bottomrule
	\end{tabular}
\end{table}

Both UV-Net and PointNext again demonstrate substantial improvements over the null model baseline across both complexity prediction tasks.
For the configurational complexity task (part flip prediction), UV-Net achieves superior performance with an MAE of 0.54 and MAPE of 35.52\%, compared to PointNext's MAE of 0.59 and MAPE of 39.33\%.
Similarly, for the geometric complexity task (unfolded bounding area prediction), UV-Net reaches an MAE of 14.60 and MAPE of 5.90\%, outperforming PointNext's MAE of 20.24 and MAPE of 8.28\%.
Both models exhibit stronger performance on geometric complexity compared to configurational complexity, mirroring the pattern observed in the feasibility tasks.

\section{Discussion}\label{sec:discussion}

This study clarifies and operationalizes the concept of manufacturability for data-driven DFM, addresses critical data limitations through synthetic data generation, and evaluates the current capabilities of deep learning models for manufacturability assessment.
In this context, we discuss the practical implications of the proposed taxonomy, the role of BenDFM and the benchmarking results, and how these contributions advance data-driven DFM.
Finally, we reflect on remaining challenges and avenues for future research.
\subsection{The taxonomy as a foundation for actionable DFM}

A central outcome of this work is a more precise and actionable definition of manufacturability. 
By distinguishing geometry-intrinsic versus configuration-dependent metrics, as well as feasibility versus complexity measures, the taxonomy clarifies what the models are predicting and the underlying assumptions driving those predictions.
This conceptual clarity has immediate practical implications. 
Models trained on geometry-intrinsic metrics can provide general, early-stage feedback within CAD environments without requiring detailed production data.
In contrast, configuration-dependent models can offer high-fidelity, context-specific insights but may not transfer well across manufacturing setups.
By explicitly framing this trade-off between generalizability and specificity, the taxonomy guides future research on transferability, domain adaptation, and cross-configuration learning.

Our work invites further exploration of hybrid metrics that combine geometric and configurational factors, as well as creative approaches of measuring the uncertainty associated with configuration-dependent predictions.
For example, composite manufacturability scores could be derived from geometric metrics to measure a notion of inherent part-complexity, whereas for configuration-dependent metrics, uncertainty quantification techniques could be employed to express confidence in predictions based on the similarity of the input geometry either to already manufactured parts in this configuration, or similarity between the current configuration and one with which the product was deemed manufacturable.
In a similar way, metrics from one quadrant could be transformed into another to provide new perspectives on manufacturability.
For instance, if manufacturability data is available across multiple configurations, multiple configurational feasibility labels could be aggregated into a configuration-independent complexity score, reflecting the proportion of configurations under which a part is manufacturable.

From a stakeholder perspective, the taxonomy enables clearer communication of manufacturability requirements across the supply chain.
By reasoning about configuration-dependency, organizations can communicate which constraints are universal (valid regardless of manufacturing partner) versus facility-specific constraints that may vary across suppliers.
This distinction streamlines supplier qualification processes and reduces iterations that result from misaligned manufacturability assumptions.
Emerging paradigms, such as federated learning, could leverage the taxonomy by collecting configuration-dependent data from multiple manufacturers and training a shared model to learn the distinct capabilities of each, all while maintaining data privacy.
This approach, recently explored in supplier selection based on process capabilities~\citep{yan2024federated}, could be effectively extended to IPMA.

\subsection{The role of BenDFM and benchmarking experiments}

BenDFM operationalizes the taxonomy and mitigates a key barrier in data-driven DFM research: the lack of advanced, representative synthetic datasets.
By capturing sequential bending operations, part-tool interactions, and both folded and unfolded geometries, BenDFM enables exploration across all four quadrants of the taxonomy.
Its process-aware generation pipeline produces realistic geometric diversity and complexity, surpasses prior datasets in fidelity and annotation richness, and provides a blueprint for creating analogous datasets in other manufacturing processes.
The global nature of many bending constraints (e.g., tool collisions, unfolding overlaps) contrasts with the local constraints typically emphasized in machining datasets (e.g., undercuts, hole depths), highlighting BenDFM's value for advancing geometric deep learning architectures capable of capturing global relationships.
Additionally, the inclusion of unfolded geometries extends its relevance to tasks such as laser cutting, nesting optimization, and multi-process DFM where cutting and bending interact.

Benchmarking experiments confirm insights from the taxonomy and highlight both the promise and limitations of current deep learning approaches. 
Topology-preserving architectures outperform point-based models, emphasizing the importance of retaining CAD-native structure for tasks sensitive to subtle geometric cues~\citep{li2025classification, hussong2025selection, zheng2025sfrgnn}. 
Configuration-dependent tasks remain challenging, illustrating the difficulty of capturing sequencing and tooling effects from geometry alone. 
The taxonomy provides a framework to reason about how much of this configurational information can be learned implicitly from labels versus requiring explicit input, balancing predictive fidelity with data requirements.
Beyond generalizability, the taxonomy encourages exploration of hybrid approaches that combine rule-based preprocessing (e.g., extracting bend angles or adjacency patterns) with learned representations, potentially improving both accuracy and interpretability.

The imperfect performance of the benchmarked models also highlights the importance of human oversight in deploying learning-based DFM tools as decision-support systems.
Since these tools often operate as black-box models, offering limited insight into how predictions are derived~\citep{lipton2018mythos}, future work should enhance model explainability in B-rep learning, extending techniques from voxel-based explainability in CAD~\citep{yoo2021explainable, ning2020manufacturing}, to ensure responsible deployment.
Furthermore, collaborative approaches that integrate model predictions with human expertise and frameworks for retrieving unstructured manufacturing knowledge during the design phase~\citep{jing2024xmkr} could enhance reliability, particularly for high-stakes manufacturability decisions~\citep{modi2011socially}.

\subsection{Limitations and future directions}

Despite these advances, several important challenges remain.
From an algorithmic perspective, bending manufacturability depends on global, long-range interactions between potentially distant bends and tooling constraints, rather than on local geometric features alone.
Current architectures, while effective at encoding local surface properties, may not fully capture these complex dependencies.
This challenge distinguishes BenDFM from existing manufacturing benchmarks, which often focus on local tasks such as machining feature recognition.
Moreover, the sequential nature of bending operations is not reflected in current model inputs, which represent only the final CAD geometry.
Sequential architectures that process the bend sequence incrementally, predicting manufacturability at each step, represent a promising direction for capturing operation-order dependencies.

BenDFM is generated using one fixed punch and die configuration, meaning current experiments primarily assess how well models can infer this implicit configuration from geometry.
Future work should investigate cross-configuration generalization by training on one tooling or machine setup and testing on another, with possible solutions using domain adaptation techniques such as few-shot learning or meta-learning to enable rapid adaptation to new configurations.
An equally relevant inverse problem is whether models can learn to recommend tooling configurations from a set of candidates, transforming configuration-dependent manufacturability assessment into a configuration-selection task.
The current binary classification of feasibility could also be extended to per-bend collision segmentation or metrics quantifying closeness to collision boundaries, providing more granular and actionable feedback.
As with all synthetic datasets, assessing how models trained on BenDFM generalize to real-world parts is a critical next step.
Fabricating a representative subset of parts would allow direct comparison between simulated labels and physical outcomes, capturing phenomena such as springback, elastic recovery, or ergonomic handling constraints not represented in the dataset.

While this work focuses on manufacturability in sheet metal bending, the underlying concepts readily extend to other forming processes such as deep drawing, stamping, and tube bending, as well as to multi-process scenarios (e.g., combined cutting and bending operations).
Importantly, the proposed taxonomy is intentionally process-agnostic and can serve as a structured foundation for dataset creation and model development across different manufacturing domains.
More broadly, our taxonomy and dataset generation methodology align with emerging design-for-X frameworks that leverage similar CAD-based learning approaches.
Recent work within Design for Assembly (DFA) has shown that generating synthetic CAD data can reduce the need for manual data collection and annotation for assembly quality inspection~\citep{zhu2025automated}.
Similarly, work within Design for Sustainability (DFS) has shown potential in estimating carbon footprint directly from CAD geometry in the design stage~\citep{hasebe2024deep}.
These efforts highlight potential synergies with our approach, suggesting that the concepts introduced here could support a broader shift toward integrated, data-driven design-for-X tools.

\section{Conclusion}\label{sec:conclusion}

We provide a practical foundation for learning-based DFM by resolving ambiguities in manufacturability definitions and addressing critical data limitations. 
Our taxonomy clarifies manufacturability along two key axes—configuration dependence and feasibility versus complexity—offering researchers and practitioners a consistent vocabulary and framework for reasoning about model applicability, generalizability, and deployment assumptions.
BenDFM operationalizes this taxonomy as a publicly released, process-aware synthetic dataset of 20,000 sheet-metal bending parts, including folded and unfolded geometries, tooling-collision labels, and continuous complexity metrics across all taxonomy quadrants.
This richness enables systematic study of previously inaccessible manufacturability phenomena and provides a foundation for researchers to benchmark geometry-aware learning approaches independently.
Benchmarking shows that topology-preserving deep learning models capture manufacturability signals more reliably than point-cloud-based models, while configuration-dependent targets remain the most challenging to learn.
Key open challenges include capturing long-range geometric dependencies inherent in bending manufacturability, generalizing across tooling configurations, and bridging the gap between synthetic benchmarks and real-world deployment.
By combining a clear taxonomy, a richly annotated dataset, and empirical insights, this work fills a major gap in manufacturability assessment and takes an important step toward industry-ready, learning-assisted DFM.

\section{Acknowledgements}
This research was supported by Flanders Make, the strategic research centre for the Flemish manufacturing industry.

\bibliography{references}

@book{anderson2020design,
  title     = {Design for manufacturability: how to use concurrent engineering to rapidly develop low-cost, high-quality products for lean production},
  author    = {Anderson, David M},
  year      = {2020},
  publisher = {Productivity Press},
  doi       = {10.4324/9780429285981}
}

@article{arinez2020artificial,
  author    = {Arinez, Jorge F and Chang, Qing and Gao, Robert X and Xu, Chengying and Zhang, Jianjing},
  journal   = {Journal of Manufacturing Science and Engineering},
  title     = {Artificial intelligence in advanced manufacturing: Current status and future outlook},
  year      = {2020},
  number    = {11},
  pages     = {110804},
  volume    = {142},
  doi       = {10.1115/1.4047855},
  publisher = {American Society of Mechanical Engineers}
}

@article{barda2023generative,
  title     = {Generative Design of Sheet Metal Structures},
  author    = {Barda, Amir and Tevet, Guy and Schulz, Adriana and Bermano, Amit Haim},
  journal   = {ACM Transactions on Graphics (TOG)},
  volume    = {42},
  number    = {4},
  pages     = {1--13},
  year      = {2023},
  publisher = {ACM New York, NY, USA},
  doi       = {10.1145/3592444}
}

@article{chen2021design,
  author    = {Chen, Niechen and Frank, Matthew C},
  journal   = {Journal of Manufacturing Science and Engineering},
  title     = {Design for manufacturing: Geometric manufacturability evaluation for five-axis milling},
  year      = {2021},
  number    = {8},
  pages     = {081007},
  volume    = {143},
  doi       = {10.1115/1.4050184},
  publisher = {American Society of Mechanical Engineers}
}

@article{chen2025virl,
  author    = {Chen, Yu-hsuan and Cagan, Jonathan and Kara, Levent Burak},
  journal   = {Journal of Intelligent Manufacturing},
  title     = {{VIRL}: Volume-informed representation learning towards few-shot manufacturability estimation},
  year      = {2025},
  pages     = {1--21},
  doi       = {10.1007/s10845-025-02575-8},
  publisher = {Springer}
}

@article{colligan2022hierarchical,
  author    = {Colligan, Andrew R and Robinson, Trevor T and Nolan, Declan C and Hua, Yang and Cao, Weijuan},
  journal   = {Computer-Aided Design},
  title     = {Hierarchical cadnet: Learning from b-reps for machining feature recognition},
  year      = {2022},
  pages     = {103226},
  volume    = {147},
  doi       = {10.1016/j.cad.2022.103226},
  publisher = {Elsevier}
}

@article{dewhurst1988early,
  title     = {Early cost estimating in product design},
  author    = {Dewhurst, P and Boothroyd, G},
  fjournal  = {Journal of Manufacturing systems},
  journal   = {J Manuf Syst},
  volume    = {7},
  number    = {3},
  pages     = {183--191},
  year      = {1988},
  publisher = {Elsevier},
  doi       = {10.1016/0278-6125(88)90003-9}
}

@inproceedings{doellken2020challenges,
  title        = {Challenges faced by design engineers when considering manufacturing in design--an interview study},
  author       = {Doellken, Markus and Zimmerer, Christoph and Matthiesen, Sven},
  booktitle    = {Proceedings of the Design Society: DESIGN Conference},
  volume       = {1},
  pages        = {837--846},
  year         = {2020},
  organization = {Cambridge University Press},
  doi = {10.1017/dsd.2020.302}
}

@article{favi2021cad,
  title     = {A {CAD}-based design for manufacturing method for casted components},
  author    = {Favi, Claudio and Mandolini, Marco and Campi, Federico and Germani, Michele},
  journal   = {Procedia CIRP},
  volume    = {100},
  pages     = {235--240},
  year      = {2021},
  publisher = {Elsevier},
  doi       = {10.1016/j.procir.2021.05.061}
}

@article{featurenet,
  author    = {Zhang, Zhibo and Jaiswal, Prakhar and Rai, Rahul},
  journal   = {Computer-Aided Design},
  title     = {Featurenet: Machining feature recognition based on 3d convolution neural network},
  year      = {2018},
  pages     = {12--22},
  volume    = {101},
  doi       = {10.1016/j.cad.2018.03.006},
  publisher = {Elsevier}
}

@article{ferrer2010methodology,
  title   = {Methodology for capturing and formalizing DFM Knowledge},
  journal = {Robotics and Computer-Integrated Manufacturing},
  volume  = {26},
  number  = {5},
  pages   = {420-429},
  year    = {2010},
  issn    = {0736-5845},
  author  = {I. Ferrer and J. Rios and J. Ciurana and M.L. Garcia-Romeu},
  doi     = {10.1016/j.rcim.2009.12.003}
}

@article{ghadai2018learning,
  author    = {Ghadai, Sambit and Balu, Aditya and Sarkar, Soumik and Krishnamurthy, Adarsh},
  title     = {Learning localized features in {3D CAD} models for manufacturability analysis of drilled holes},
  pages     = {263--275},
  volume    = {62},
  fjournal  = {Computer Aided Geometric Design},
  journal   = {Comput Aided Geom D},
  publisher = {Elsevier},
  year      = {2018},
  doi       = {10.1016/j.cagd.2018.03.024}
}

@inproceedings{girsule2020data,
  title        = {Data Acquisition Approaches for {AI}-supported Metal Processing},
  author       = {Girsule, Bernhard and Rottermanner, Gernot and Jandl, Christian and Kreiger, Mylene and Moser, Thomas and Fuchs, Patricia},
  booktitle    = {2020 25th IEEE International Conference on Emerging Technologies and Factory Automation (ETFA)},
  volume       = {1},
  pages        = {1027--1030},
  year         = {2020},
  organization = {IEEE},
  doi          = {10.1109/ETFA46521.2020.9211935}
}

@article{gupta1997automated,
  author    = {Gupta, Satyandra K and Regli, William C and Das, Diganta and Nau, Dana S},
  journal   = {Research in Engineering Design},
  title     = {Automated manufacturability analysis: A survey},
  year      = {1997},
  pages     = {168--190},
  volume    = {9},
  doi       = {10.1007/BF01596601},
  publisher = {Springer}
}

@article{gupta2013classification,
  author    = {Gupta, Ravi Kumar and Gurumoorthy, Balan},
  journal   = {Computer-Aided Design},
  title     = {Classification, representation, and automatic extraction of deformation features in sheet metal parts},
  year      = {2013},
  number    = {11},
  pages     = {1469--1484},
  volume    = {45},
  doi       = {10.1016/j.cad.2013.06.010},
  publisher = {Elsevier}
}

@article{hamouche2018classification,
  author    = {Hamouche, Elia and Loukaides, Evripides G},
  journal   = {International Journal of Computer Integrated Manufacturing},
  title     = {Classification and selection of sheet forming processes with machine learning},
  year      = {2018},
  number    = {9},
  pages     = {921--932},
  volume    = {31},
  doi       = {10.1080/0951192X.2018.1429668},
  publisher = {Taylor \& Francis}
}

@article{hasebe2024deep,
  title     = {Deep CAD Shape Recognition for Carbon Footprint Estimation at the Design Stage},
  author    = {Hasebe, Tatsuya and Katayama, Erika and Yoshiteru, Katsumura},
  journal   = {Procedia CIRP},
  volume    = {122},
  pages     = {545--550},
  year      = {2024},
  publisher = {Elsevier},
  doi = {10.1016/j.procir.2024.01.080}
}

@inproceedings{henderson2018deep,
  title     = {Deep reinforcement learning that matters},
  author    = {Henderson, Peter and Islam, Riashat and Bachman, Philip and Pineau, Joelle and Precup, Doina and Meger, David},
  booktitle = {Proceedings of the AAAI conference on artificial intelligence},
  volume    = {32},
  number    = {1},
  year      = {2018},
  doi       = {10.5555/3504035.3504427}
}

@article{hoefer2018automated,
  author    = {Hoefer, Michael J and Frank, Matthew C},
  journal   = {Journal of Mechanical Design},
  title     = {Automated manufacturing process selection during conceptual design},
  year      = {2018},
  number    = {3},
  pages     = {031701},
  volume    = {140},
  doi       = {10.1115/1.4038686},
  publisher = {American Society of Mechanical Engineers}
}

@article{hussong2025selection,
  author    = {Hussong, Marco and Ruediger-Flore, Patrick and Klar, Matthias and Kloft, Marius and Aurich, Jan C},
  journal   = {Journal of Manufacturing Systems},
  title     = {Selection of manufacturing processes using graph neural networks},
  year      = {2025},
  pages     = {176--193},
  volume    = {80},
  doi       = {10.1016/j.jmsy.2025.02.016},
  publisher = {Elsevier}
}

@inproceedings{jayaraman2021uv,
  author    = {Jayaraman, Pradeep Kumar and Sanghi, Aditya and Lambourne, Joseph G and Willis, Karl DD and Davies, Thomas and Shayani, Hooman and Morris, Nigel},
  booktitle = {Proceedings of the IEEE/CVF Conference on Computer Vision and Pattern Recognition (CVPR)},
  title     = {Uv-net: Learning from boundary representations},
  pages     = {11703--11712},
  year      = {2021},
  doi       = {10.1109/CVPR46437.2021.01153}
}

@article{jing2024xmkr,
  title     = {XMKR: Explainable manufacturing knowledge recommendation for collaborative design with graph embedding learning},
  author    = {Jing, Yanzhen and Zhou, Guanghui and Zhang, Chao and Chang, Fengtian and Yan, Hairui and Xiao, Zhongdong},
  journal   = {Advanced Engineering Informatics},
  volume    = {59},
  pages     = {102339},
  year      = {2024},
  publisher = {Elsevier},
  doi = {10.1016/j.aei.2023.102339}
}

@incollection{kumar2016feature,
  title     = {Feature extraction and manufacturability assessment of sheet metal parts},
  author    = {Kumar, Shailendra and Singh, Rajender and Panghal, Deepak and Salunkhe, Sachin and Hussein, Hussein MA},
  booktitle = {AI Applications in Sheet Metal Forming},
  pages     = {41--66},
  year      = {2016},
  publisher = {Springer},
  doi       = {10.1007/978-981-10-2251-7_3}
}

@article{lehrer2025uscm,
  title   = {UCSM: Dataset of U-shaped parametric CAD geometries and real-world sheet metal meshes for deep drawing},
  journal = {Computer-Aided Design},
  pages   = {103924},
  year    = {2025},
  issn    = {0010-4485},
  author  = {Tobias Lehrer and Philipp Stocker and Fabian Duddeck and Marcus Wagner},
  doi     = {10.1016/j.cad.2025.103924}
}

@article{li2025classification,
  author    = {Li, Zirui and Tong, Xiaomeng and Shi, Peng and Cai, Maolin and Han, Fei},
  journal   = {Engineering Applications of Artificial Intelligence},
  title     = {Classification of non-standard mechanical parts with graph convolutional networks},
  year      = {2025},
  pages     = {110879},
  volume    = {153},
  doi       = {10.1016/j.engappai.2025.110879},
  publisher = {Elsevier}
}

@article{liu2025manufacturing,
  title     = {Manufacturing process identification from 3D point cloud models using semantic segmentation},
  author    = {Liu, Xiaofang and Wang, Zhichao and Melkote, Shreyes N and Rosen, David W},
  journal   = {Journal of Manufacturing Systems},
  volume    = {82},
  pages     = {858--873},
  year      = {2025},
  publisher = {Elsevier},
  doi       = {10.1016/j.jmsy.2025.07.023}
}

@inproceedings{mfcad,
  author       = {Cao, Weijuan and Robinson, Trevor and Hua, Yang and Boussuge, Flavien and Colligan, Andrew R and Pan, Wanbin},
  booktitle    = {International Design Engineering Technical Conferences \& Computers and Information in Engineering Conference},
  title        = {Graph representation of {3D CAD} models for machining feature recognition with deep learning},
  year         = {2020},
  organization = {American Society of Mechanical Engineers},
  pages        = {V11AT11A003},
  volume       = {84003},
  doi          = {10.1115/DETC2020-22355}
}

@article{modi2011socially,
  title     = {A socially inspired framework for human state inference using expert opinion integration},
  author    = {Modi, Shrey and Lin, Yingzi and Cheng, Long and Yang, Guosheng and Liu, Lizhi and Zhang, WJ},
  journal   = {IEEE/ASME Transactions on Mechatronics},
  volume    = {16},
  number    = {5},
  pages     = {874--878},
  year      = {2011},
  publisher = {IEEE},
  doi = {10.1109/TMECH.2011.2161094}
}

@article{naranje2011knowledge,
  author    = {Naranje, Vishal and Kumar, Shailendra},
  journal   = {Key Engineering Materials},
  title     = {A knowledge-based system for manufacturability assessment of deep drawn sheet metal parts},
  year      = {2011},
  pages     = {749--756},
  volume    = {473},
  doi       = {10.4028/www.scientific.net/KEM.473.749},
  publisher = {Trans Tech Publ}
}

@article{ning2020manufacturing,
  author    = {Ning, Fangwei and Shi, Yan and Cai, Maolin and Xu, Weiqing and Zhang, Xianzhi},
  title     = {Manufacturing cost estimation based on a deep-learning method},
  pages     = {186--195},
  volume    = {54},
  fjournal  = {Journal of Manufacturing Systems},
  journal   = {J Manuf Syst},
  publisher = {Elsevier},
  year      = {2020},
  doi       = {10.1016/j.jmsy.2019.12.005}
}

@article{ning2020manufacturing2,
  author    = {Ning, Fangwei and Shi, Yan and Cai, Maolin and Xu, Weiqing and Zhang, Xianzhi},
  journal   = {Journal of Manufacturing Systems},
  title     = {Manufacturing cost estimation based on the machining process and deep-learning method},
  year      = {2020},
  pages     = {11--22},
  volume    = {56},
  doi       = {10.1016/j.jmsy.2020.04.011},
  publisher = {Elsevier}
}

@article{ning2023part,
  author    = {Ning, Fangwei and Shi, Yan and Cai, Maolin and Xu, Weiqing},
  journal   = {Journal of Intelligent Manufacturing},
  title     = {Part machining feature recognition based on a deep learning method},
  year      = {2023},
  number    = {2},
  pages     = {809--821},
  volume    = {34},
  doi       = {10.1007/s10845-021-01827-7},
  publisher = {Springer}
}

@inproceedings{niu2024manufacturing,
  title        = {Manufacturing Process Selection Using Graph Neural Networks: A Novel Approach to {CAD} Model Semantic Segmentation},
  author       = {Niu, Shuai and Cai, Yishu and Li, Feng and Li, Yibo and Tong, Xiaomeng},
  booktitle    = {2024 4th International Symposium on Artificial Intelligence and Intelligent Manufacturing (AIIM)},
  pages        = {124--129},
  year         = {2024},
  organization = {IEEE},
  doi          = {10.1109/AIIM64537.2024.10934313}
}

@article{ordek2024machine,
  author    = {{\"O}rdek, Baris and Borgianni, Yuri and Coatanea, Eric},
  journal   = {Production \& Manufacturing Research},
  title     = {Machine learning-supported manufacturing: A review and directions for future research},
  year      = {2024},
  number    = {1},
  pages     = {2326526},
  volume    = {12},
  doi       = {10.1080/21693277.2024.2326526},
  publisher = {Taylor \& Francis}
}

@article{lipton2018mythos,
  title={The mythos of model interpretability: In machine learning, the concept of interpretability is both important and slippery.},
  author={Lipton, Zachary C},
  journal={Queue},
  volume={16},
  number={3},
  pages={31--57},
  year={2018},
  publisher={ACM New York, NY, USA},
  doi={10.1145/3236386.3241340}
}

@article{patterson2021generation,
  title     = {Generation and enforcement of process-driven manufacturability constraints: A survey of methods and perspectives for product design},
  author    = {Patterson, Albert E and Lee, Yong Hoon and Allison, James T},
  journal   = {Journal of Mechanical Design},
  volume    = {143},
  number    = {11},
  pages     = {110801},
  year      = {2021},
  publisher = {American Society of Mechanical Engineers},
  doi = {10.1115/1.4050740}
}

@misc{paviot2022pythonocc,
  author    = {Thomas Paviot},
  title     = {pythonocc},
  version   = {7.7.0},
  year      = {2022},
  publisher = {Zenodo},
  doi       = {10.5281/zenodo.7471333}
}

@article{peddireddy2021identifying,
  author    = {Peddireddy, Dheeraj and Fu, Xingyu and Shankar, Anirudh and Wang, Haobo and Joung, Byung Gun and Aggarwal, Vaneet and Sutherland, John W and Jun, Martin Byung-Guk},
  journal   = {Journal of Manufacturing Processes},
  title     = {Identifying manufacturability and machining processes using deep 3{D} convolutional networks},
  year      = {2021},
  pages     = {1336--1348},
  volume    = {64},
  doi       = {10.1016/j.jmapro.2021.02.034},
  publisher = {Elsevier}
}

@article{pullan2010application,
  title     = {Application of concurrent engineering in manufacturing industry},
  author    = {Pullan, Thankachan T and Bhasi, M and Madhu, G},
  journal   = {International Journal of Computer Integrated Manufacturing},
  volume    = {23},
  number    = {5},
  pages     = {425--440},
  year      = {2010},
  publisher = {Taylor \& Francis},
  doi       = {10.1080/09511921003643152}
}

@article{qian2022pointnext,
  author  = {Qian, Guocheng and Li, Yuchen and Peng, Houwen and Mai, Jinjie and Hammoud, Hasan and Elhoseiny, Mohamed and Ghanem, Bernard},
  journal = {Advances in Neural Information Processing Systems},
  title   = {Pointnext: Revisiting pointnet++ with improved training and scaling strategies},
  year    = {2022},
  pages   = {23192--23204},
  volume  = {35},
  doi     = {10.5555/3600270.3601955}
}

@article{salem2017towards,
  author    = {Salem, Amr A and Abdelmaguid, Tamer F and Wifi, Abdalla S and Elmokadem, Alaa},
  journal   = {International Journal of Advanced Manufacturing Technology},
  title     = {Towards an efficient process planning of the {V}-bending process: an enhanced automated feature recognition system},
  year      = {2017},
  pages     = {4163--4181},
  volume    = {91},
  doi       = {10.1007/s00170-017-0104-9},
  publisher = {Springer}
}

@article{seibold2022process,
  author    = {Seibold, Johannes and Hentsch, Maximilian and Kharitonov, Aleksei and Eber, Rainer and Schwarzer, Steffen},
  journal   = {Procedia Computer Science},
  title     = {Process automation in the area of manufacturability analysis using machine learning},
  year      = {2022},
  pages     = {196--204},
  volume    = {204},
  doi       = {10.1016/j.procs.2022.08.024},
  publisher = {Elsevier}
}

@incollection{shankar1993generalized,
  title     = {A generalized methodology for evaluating manufacturability},
  author    = {Shankar, Srinivasa R and Jansson, David G},
  booktitle = {Concurrent Engineering: Contemporary issues and modern design tools},
  pages     = {248--263},
  year      = {1993},
  publisher = {Springer},
  doi       = {10.1007/978-1-4615-3062-6_13}
}

@article{sharp2018survey,
  author    = {Sharp, Michael and Ak, Ronay and Hedberg Jr, Thomas},
  journal   = {Journal of Manufacturing Systems},
  title     = {A survey of the advancing use and development of machine learning in smart manufacturing},
  year      = {2018},
  pages     = {170--179},
  volume    = {48},
  doi       = {10.1016/j.jmsy.2018.02.004},
  publisher = {Elsevier}
}

@article{smcad,
  author    = {Ma, Liuhuan and Yang, Jiong},
  journal   = {Scientific Reports},
  title     = {Adaptive recognition of machining features in sheet metal parts based on a graph class-incremental learning strategy},
  year      = {2024},
  number    = {1},
  pages     = {10656},
  volume    = {14},
  doi       = {10.1038/s41598-024-61443-2},
  publisher = {Nature Publishing Group UK London}
}

@article{stjepandic2015intellectual,
  title     = {Intellectual property protection},
  author    = {Stjepandi{\'c}, Josip and Liese, Harald and Trappey, Amy JC},
  journal   = {Concurrent engineering in the 21st century: Foundations, developments and challenges},
  pages     = {521--551},
  year      = {2015},
  publisher = {Springer},
  doi       = {10.1007/978-3-319-13776-6}
}

@book{swift2013manufacturing,
  title     = {Manufacturing process selection handbook},
  author    = {Swift, KG and Booker, JD},
  year      = {2013},
  publisher = {Butterworth-Heinemann},
  doi       = {10.1016/C2011-0-07343-X}
}

@article{wang2018deep,
  author    = {Wang, Jinjiang and Ma, Yulin and Zhang, Laibin and Gao, Robert X and Wu, Dazhong},
  journal   = {Journal of Manufacturing Systems},
  title     = {Deep learning for smart manufacturing: Methods and applications},
  year      = {2018},
  pages     = {144--156},
  volume    = {48},
  doi       = {10.1016/j.jmsy.2018.01.003},
  publisher = {Elsevier}
}

@article{wang2022machine,
  author    = {Wang, Xingzhi and Liu, Ang and Kara, Sami},
  journal   = {J Manuf Syst},
  title     = {Machine learning for engineering design toward smart customization: A systematic review},
  year      = {2022},
  pages     = {391--405},
  volume    = {65},
  fjournal  = {Journal of Manufacturing Systems},
  publisher = {Elsevier},
  doi       = {10.1016/j.jmsy.2022.10.001}
}

@article{wang2023manufacturing,
  author    = {Wang, Zhichao and Rosen, David},
  journal   = {Journal of Intelligent Manufacturing},
  title     = {Manufacturing process classification based on heat kernel signature and convolutional neural networks},
  year      = {2023},
  number    = {8},
  pages     = {3389--3411},
  volume    = {34},
  doi       = {10.1007/s10845-022-02009-9},
  publisher = {Springer}
}

@article{wang2023manufacturing2,
  author    = {Wang, Zhichao and Rosen, David},
  journal   = {Journal of Computing and Information Science in Engineering},
  title     = {Manufacturing process classification based on distance rotationally invariant convolutions},
  year      = {2023},
  number    = {5},
  pages     = {051004},
  volume    = {23},
  doi       = {10.1115/1.4056806},
  publisher = {American Society of Mechanical Engineers}
}

@article{xiao2025novel,
  author    = {Xiao, Biao and Zhao, Zhengcai and Xu, Baode and Li, Yao and Zhang, Wei and Huan, Haixiang and Su, Honghua},
  journal   = {Journal of Manufacturing Systems},
  title     = {A novel method for intelligent reasoning of machining step sequences based on deep reinforcement learning},
  year      = {2025},
  pages     = {626--642},
  volume    = {80},
  doi       = {10.1016/j.jmsy.2025.04.005},
  publisher = {Elsevier}
}

@article{yan2023automated,
  author    = {Yan, Xiaoliang and Melkote, Shreyes},
  journal   = {Journal of Manufacturing Systems},
  title     = {Automated manufacturability analysis and machining process selection using deep generative model and Siamese neural networks},
  year      = {2023},
  pages     = {57--67},
  volume    = {67},
  doi       = {10.1016/j.jmsy.2023.01.006},
  publisher = {Elsevier}
}

@article{yan2024deep,
  author    = {Yan, Xiaoliang and Williams, Reed and Arvanitis, Elena and Melkote, Shreyes},
  journal   = {Journal of Manufacturing Systems},
  title     = {Deep learning-based semantic segmentation of machinable volumes for cyber manufacturing service},
  year      = {2024},
  pages     = {16--25},
  volume    = {72},
  doi       = {10.1016/j.jmsy.2023.11.005},
  publisher = {Elsevier}
}

@article{yan2024federated,
  author    = {Yan, Xiaoliang and Wang, Zhichao and Puvvada, Mukunda Moulik and Dinar, Mahmoud and Rosen, David W and Melkote, Shreyes N},
  journal   = {J Manuf Syst},
  title     = {A federated learning approach to automated and secure supplier selection in cyber manufacturing as-a-service},
  year      = {2024},
  pages     = {170--183},
  volume    = {77},
  fjournal  = {Journal of Manufacturing Systems},
  publisher = {Elsevier},
  doi       = {10.1016/j.jmsy.2024.09.005}
}

@article{yeo2021manufacturability,
  author    = {Yeo, Changmo and Cheon, Sanguk and Mun, Duhwan},
  journal   = {International Journal of Computer Integrated Manufacturing},
  title     = {Manufacturability evaluation of parts using descriptor-based machining feature recognition},
  year      = {2021},
  number    = {11},
  pages     = {1196--1222},
  volume    = {34},
  doi       = {10.1080/0951192X.2021.1963483},
  publisher = {Taylor \& Francis}
}

@article{yoo2021explainable,
  title     = {Explainable artificial intelligence for manufacturing cost estimation and machining feature visualization},
  author    = {Yoo, Soyoung and Kang, Namwoo},
  journal   = {Expert Systems with Applications},
  volume    = {183},
  pages     = {115430},
  year      = {2021},
  publisher = {Elsevier}
}

@article{yoo2021integrating,
  author    = {Yoo, Soyoung and Lee, Sunghee and Kim, Seongsin and Hwang, Kwang Hyeon and Park, Jong Ho and Kang, Namwoo},
  journal   = {Structural and Multidisciplinary Optimization},
  title     = {Integrating deep learning into CAD/CAE system: generative design and evaluation of 3D conceptual wheel},
  year      = {2021},
  number    = {4},
  pages     = {2725--2747},
  volume    = {64},
  doi       = {10.1007/s00158-021-02953-9},
  publisher = {Springer}
}

@article{zhang2022machining,
  author    = {Zhang, Hang and Zhang, Shusheng and Zhang, Yajun and Liang, Jiachen and Wang, Zhen},
  journal   = {Robotics and Computer-Integrated Manufacturing},
  title     = {Machining feature recognition based on a novel multi-task deep learning network},
  year      = {2022},
  pages     = {102369},
  volume    = {77},
  doi       = {10.1016/j.rcim.2022.102369},
  publisher = {Elsevier}
}

@article{zhang2022novel,
  author    = {Zhang, Hang and Wang, Wenhu and Zhang, Shusheng and Huang, Bo and Zhang, Yajun and Wang, Mingwei and Liang, Jiachen and Wang, Zhen},
  journal   = {Journal of Manufacturing Systems},
  title     = {A novel method based on a convolutional graph neural network for manufacturing cost estimation},
  year      = {2022},
  pages     = {837--852},
  volume    = {65},
  doi       = {10.1016/j.jmsy.2022.10.007},
  publisher = {Elsevier}
}

@article{zhang2024brepmfr,
  author    = {Zhang, Shuming and Guan, Zhidong and Jiang, Hao and Wang, Xiaodong and Tan, Pingan},
  journal   = {Computer Aided Geometric Design},
  title     = {BrepMFR: Enhancing machining feature recognition in B-rep models through deep learning and domain adaptation},
  year      = {2024},
  pages     = {102318},
  volume    = {111},
  doi       = {10.1016/j.cagd.2024.102318},
  publisher = {Elsevier}
}

@article{zhang2024point,
  author    = {Zhang, Hang and Wang, Wenhu and Zhang, Shusheng and Wang, Zhen and Zhang, Yajun and Zhou, Jingtao and Huang, Bo},
  journal   = {Journal of Manufacturing Systems},
  title     = {Point cloud self-supervised learning for machining feature recognition},
  year      = {2024},
  pages     = {78--95},
  volume    = {77},
  doi       = {10.1016/j.jmsy.2024.08.029},
  publisher = {Elsevier}
}

@article{zhao2020automated,
  author    = {Zhao, Changxuan and Dinar, Mahmoud and Melkote, Shreyes N},
  title     = {Automated classification of manufacturing process capability utilizing part shape, material, and quality attributes},
  number    = {2},
  pages     = {021011},
  volume    = {20},
  fjournal  = {Journal of Computing and Information Science in Engineering},
  journal   = {J Comput Inf Sci Eng},
  publisher = {American Society of Mechanical Engineers},
  year      = {2020},
  doi       = {10.1115/1.4045410}
}

@article{zhao2022data,
  author    = {Zhao, Changxuan and Dinar, Mahmoud and Melkote, Shreyes N},
  journal   = {Journal of Manufacturing Systems},
  title     = {A data-driven framework for learning the capability of manufacturing process sequences},
  year      = {2022},
  pages     = {68--80},
  volume    = {64},
  doi       = {10.1016/j.jmsy.2022.05.009},
  publisher = {Elsevier}
}

@article{zhao2024deep,
  author    = {Zhao, Changxuan and Dinar, Mahmoud and Melkote, Shreyes N},
  journal   = {International Journal of Production Research},
  title     = {Deep learning and sequence mining for manufacturing process and sequence selection},
  year      = {2024},
  number    = {14},
  pages     = {5293--5314},
  volume    = {62},
  doi       = {10.1080/00207543.2023.2290700},
  publisher = {Taylor \& Francis}
}

@article{zhao2024learning,
  author    = {Zhao, Changxuan and Melkote, Shreyes N},
  title     = {Learning the manufacturing capabilities of machining and finishing processes using a deep neural network model},
  number    = {4},
  pages     = {1845--1865},
  volume    = {35},
  fjournal  = {Journal of Intelligent Manufacturing},
  journal   = {J Intell Manuf},
  publisher = {Springer},
  year      = {2024},
  doi       = {10.1007/s10845-023-02134-z}
}

@article{zheng2025sfrgnn,
  author    = {Zheng, Xiaohu and Chen, Hongbo and He, Fangzhou and Liu, Xiaojia},
  title     = {{SFRGNN-DA}: An enhanced graph neural network with domain adaptation for feature recognition in structural parts machining},
  pages     = {16--33},
  volume    = {81},
  fjournal  = {Journal of Manufacturing Systems},
  journal   = {J Manuf Syst},
  publisher = {Elsevier},
  year      = {2025},
  doi       = {10.1016/j.jmsy.2025.05.005}
}

@inproceedings{zhong2025deepmill,
  title     = {DeepMill: Neural Accessibility Learning for Subtractive Manufacturing},
  author    = {Zhong, Fanchao and Wang, Yang and Wang, Peng-Shuai and Lu, Lin and Zhao, Haisen},
  booktitle = {Proceedings of the Special Interest Group on Computer Graphics and Interactive Techniques Conference Conference Papers},
  pages     = {1--11},
  year      = {2025},
  publisher = {ACM New York, NY, USA},
  doi       = {10.1145/3721238.3730657}
}

@article{zhu2025automated,
  title     = {Automated assembly quality inspection by deep learning with 2D and 3D synthetic CAD data},
  author    = {Zhu, Xiaomeng and M{\aa}rtensson, P{\"a}r and Hanson, Lars and Bj{\"o}rkman, M{\aa}rten and Maki, Atsuto},
  journal   = {Journal of Intelligent Manufacturing},
  volume    = {36},
  number    = {4},
  pages     = {2567--2582},
  year      = {2025},
  publisher = {Springer},
  doi={10.1007/s10845-024-02375-6}
}

\end{document}